\documentclass[lettersize,journal]{IEEEtran}
\usepackage{cite}
\usepackage{amsmath,amssymb,amsfonts}
\usepackage{graphicx}
\usepackage{textcomp}
\usepackage{algorithmic}

\usepackage{algorithm}
\usepackage{array}
\usepackage[caption=false,font=normalsize,labelfont=sf,textfont=sf]{subfig}
\usepackage{verbatim}
\usepackage{latexsym}
\usepackage{amsthm}
\usepackage{booktabs}
\usepackage{tabularx}
\usepackage{enumitem}
\usepackage{diagbox}
\usepackage{url}
\usepackage{float}
\usepackage{multirow}
\usepackage{stfloats}
\usepackage[small]{caption}
\usepackage{subcaption}
\usepackage{color}

\usepackage[morefloats=100]{morefloats}

\usepackage{highlight}

\usepackage{colortbl}
\usepackage{siunitx}
\usepackage[breaklinks]{hyperref}
\newcolumntype{L}[1]{>{\raggedright\arraybackslash}p{#1}}
\newcolumntype{C}[1]{>{\centering\arraybackslash}p{#1}}
\newcolumntype{R}[1]{>{\raggedleft\arraybackslash}p{#1}}

\newcommand{\BibTeX}{B\kern-.05em{\sc i\kern-.025em b}\kern-.08em\TeX}

\usepackage{soul}

\begin{document}


\title{The Impact of Bootstrap Sampling Rate on Random Forest Performance in Regression Tasks}
\author{Michał Iwaniuk, Mateusz Jarosz, Bartłomiej Borycki, Bartosz Jezierski, \\Jan Cwalina, Stanis{\l}aw Ka{\'z}mierczak, Jacek Ma{\'n}dziuk
\thanks{All authors are with the Faculty of Mathematics and Information Science, Warsaw University of Technology, Warsaw, Poland. Corresponding author: Stanis{\l}aw Ka{\'z}mierczak (email: stanislaw.kazmierczak@pw.edu.pl).}}

\maketitle



\begin{figure*}[b]
\centering
\setlength{\fboxsep}{8pt}
\setlength{\fboxrule}{0.5pt}
\fbox{
    \parbox{0.97\textwidth}{
        {\color{red}
        This work has been submitted to the IEEE for possible publication. 
        Copyright may be transferred without notice, after which this version 
        may no longer be accessible.}
    }
}
\end{figure*}

\begin{abstract}
Random Forests (RFs) typically train each tree on a bootstrap sample of the same size as the training set, i.e., bootstrap rate (BR) equals 1.0. We systematically examine how varying BR from 0.2 to 5.0 affects RF performance across 39 heterogeneous regression datasets and 16 RF configurations, evaluating with repeated two-fold cross-validation and mean squared error. Our results demonstrate that tuning the BR can yield significant improvements over the default: the best setup relied on BR~$\leq~$1.0 for 24 datasets, BR~$>$~1.0 for 15, and BR~$=$~1.0 was optimal in 4 cases only. We establish a link between dataset characteristics and the preferred BR: datasets with strong global feature-target relationships favor higher BRs, while those with higher local target variance benefit from lower BRs. To further investigate this relationship, we conducted experiments on synthetic datasets with controlled noise levels. These experiments reproduce the observed bias–variance trade-off: in low-noise scenarios, higher BRs effectively reduce model bias, whereas in high-noise settings, lower BRs help reduce model variance. Overall, BR is an influential hyperparameter that should be tuned to optimize RF regression models.
\end{abstract}



\section{Introduction}
\label{sec:Introduction}
\IEEEPARstart{R}{andom} Forest (RF) is an ensemble machine learning (ML) algorithm involving a set of decision trees that collectively make a decision. In classification tasks, each tree votes for a particular class, and the predicted label is determined either by hard voting (majority vote) or soft voting (averaged class probabilities across the trees). In regression tasks, the final prediction is the mean of all individual tree outputs. RFs serve as a robust baseline across a wide range of ML problems, offering 
an effective balance of predictive accuracy, training speed, and moderate interpretability. While gradient-boosted trees or deep neural networks may outperform them in heavily tuned or domain-specific settings, RF models consistently deliver near-optimal results with minimal tuning, especially on structured, tabular datasets~\cite{grinsztajn2022why,Vysotska2025111}.

The performance of an ensemble model depends primarily on two factors: the strength of the individual base models and their diversity~\cite{hastie2009elements}. In RFs, diversity arises from the randomness introduced during training, which has two main sources: the random selection of candidate features at each node split and the bootstrapping of the training data. Bootstrapping 
consists in randomly sampling observations from the original training set with replacement, thereby creating an individual training subset for each tree in the ensemble. The ratio of the number of observations in the bootstrap sample to the number of observations in the full training set is referred to as the bootstrap rate (BR). In the literature, this hyperparameter is also known as sampling rate, bag fraction, or bootstrap size ratio, among others. This study investigates the influence of the BR hyperparameter on model performance—an aspect that, as discussed in Section~\ref{sec:RelatedLiterature}, has received limited attention in the existing literature.

\begin{figure}[t]
  \centering
  \includegraphics[width=1.00\linewidth]{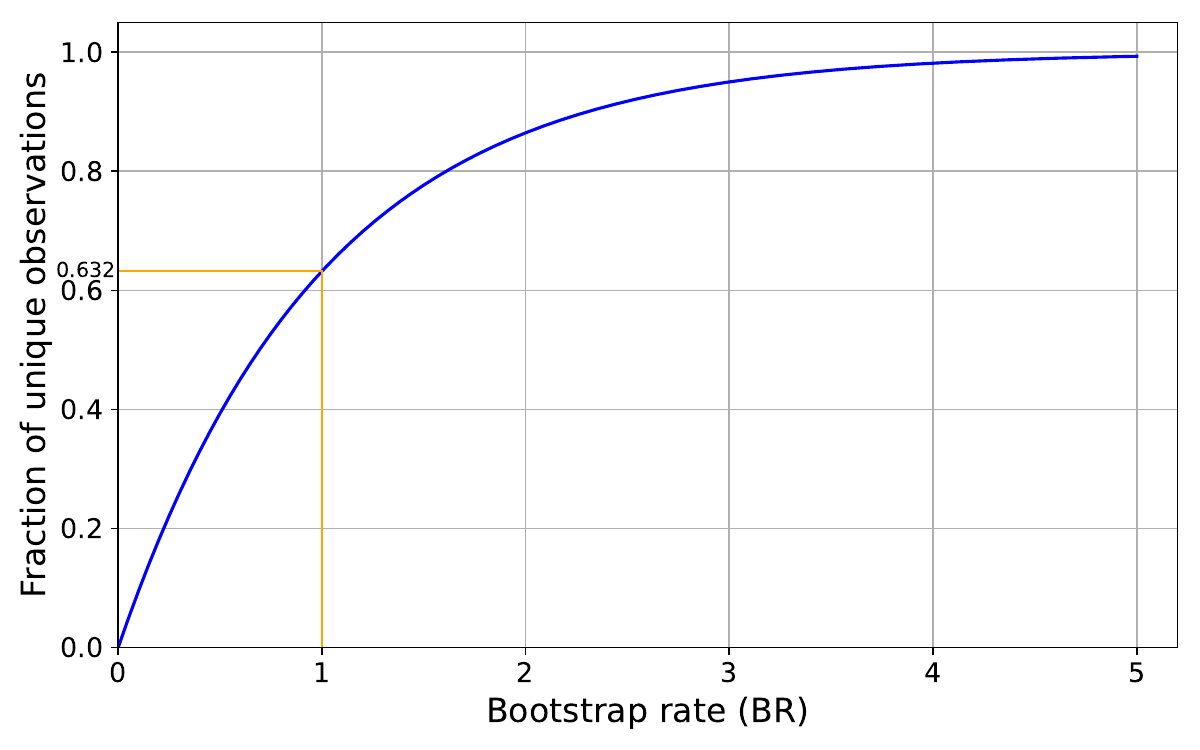}
\caption{Expected fraction of original instances that appear at least once in a bootstrap sample of size~$BR \cdot N$, drawn with replacement from a dataset of size~$N$. The orange line indicates the classical bootstrap setting with $BR~=~1.0$, where approximately 63.2\% of the original instances are expected to be included. The curve follows the approximation $\mathbb{E}(U)/N \approx 1 - e^{-BR}$, which becomes asymptotically exact as $N \to \infty$ with fixed BR.}
\label{fig:expected-distinct}
\end{figure}

BR influences the frequency with which individual observations appear in a bootstrap sample, thereby affecting the number of distinct original instances used to train each tree in an RF. As shown in~\cite{Wasserman2004}, when sampling $n$ times with replacement from a dataset of $N$ unique examples, the expected number of distinct observations in the sample $\mathbb{E}(U)$ is given by
\begin{align}
\mathbb{E}(U) &= N \cdot \left(1 - \left(1 - \frac{1}{N} \right)^n \right) \\
&= N \cdot \left(1 - \left(\left(1 - \frac{1}{N} \right)^N \right)^{BR} \right).
\end{align}
As $N \to \infty$, the formula simplifies to:
\begin{equation}
\lim_{N \to \infty} \mathbb{E}(U) = N \cdot \left(1 - e^{-BR} \right),
\label{eq:limit}
\end{equation}
since $\left(1 - \frac{1}{N} \right)^N \to e^{-1}$ as $N \to \infty$.
Fig.~\ref{fig:expected-distinct} illustrates how the expected fraction of distinct observations changes as a function of BR under sampling with replacement. In the classical bootstrap setting, where $n = N$ (i.e., BR~$=$~1.0), Eq.~\eqref{eq:limit} simplifies to
\begin{equation}
\mathbb{E}(U) \approx N \cdot \left(1 - e^{-1} \right) \approx 0.632 \cdot N.
\end{equation}
This implies that for sufficiently large $N$ (in practice, a few hundred samples suffice), a bootstrap sample contains approximately 63.2\% distinct instances from the original dataset, while the remaining examples are not included.

In this work, we aim to fill a research gap concerning the impact of the hyperparameter BR on the performance of RF models in regression tasks. Additionally, we identify the dataset characteristics that influence the optimal BR value. As shown in Fig.~\ref{fig:expected-distinct}, the expected proportion of distinct observations increases smoothly with higher BR values. Therefore, from a statistical standpoint, BR~$=$~1.0 does not represent any unique or critical threshold. Its significance is purely psychological. With this in mind, we extend our analysis to include values of BR greater than 1.0.

The main contributions of this work are as follows:
\begin{itemize}
    \item To the best of our knowledge, this work is the first to conduct a comprehensive analysis of the impact of the BR hyperparameter on the RF performance in regression tasks.
    \item We show that BR values greater than 1.0, previously not analyzed in the literature, quite often yield better results than the classical values ($\le~1.0$). These findings suggest that BR $>1.0$ should be considered during RF tuning.
    \item Similarly, we show that small BR values ($\leq 0.2$), which are rarely explored in the literature, can also lead to improved performance.
    \item We observe that the distribution of optimal BR values is highly dataset-dependent.
    \item Consequently, we investigate the relationship between the characteristics of the dataset and the suitable BR values, showing which dataset properties are relevant for the BR selection in regression tasks.
\end{itemize}

The remainder of this paper is organized as follows. Section~\ref{sec:RelatedLiterature} reviews related work on hyperparameter optimization and Random Forest tuning, with emphasis on studies addressing bootstrap sampling strategies. Section~\ref{sec:ExperimentalSetup} describes the experimental setup, including dataset preprocessing, model configurations, and evaluation protocol. Section~\ref{sec:Results} presents the results and provides a statistical analysis of the influence of the BR. Section~\ref{sec:Discussion} discusses the observed phenomena, linking optimal BR values to data characteristics and model behavior, and examines the bias--variance mechanisms using synthetic experiments. Finally, Section~\ref{sec:Conclusions} concludes the paper and outlines directions for future research.

\section{Related Literature}
\label{sec:RelatedLiterature}
Hyperparameter optimization (HPO) is a decisive factor in determining both the effectiveness and efficiency of contemporary ML methods. Recent research trends indicate a shift away from exhaustive search procedures toward data- and knowledge-driven approaches, including meta-learning and zero- or few-shot hyperparameter transfer, as well as more rigorous AutoML benchmarking. These developments reflect an increasing emphasis not only on predictive accuracy but also on computational efficiency. In this section, we review recent advances in hyperparameter optimization and AutoML, and then narrow the scope to RFs, highlighting the limited attention paid to the BR hyperparameter.

\subsection{Hyperparameter Optimization and AutoML}
A popular and foundational strategy in HPO is \emph{Sequential Model-Based Optimization} (SMBO)~\cite{10.1007/978-3-642-25566-3_40}, wherein a probabilistic surrogate is iteratively refined to balance exploration of new hyperparameter configurations with exploitation of known high-performing regions. Canonical SMBO instances include Gaussian process-based Bayesian optimization (e.g.,~\cite{snoek2012practicalbayesianoptimizationmachine}) and the Tree-structured Parzen Estimator (TPE)~\cite{10.5555/2986459.2986743}. Building on this line, Sieradzki and Ma\'{n}dziuk~\cite{Sieradzki2025EATPE} introduced EATPE, which extends TPE as well as an advanced TPE variant proposed by Arsenault~\cite{arsenault2023learning}.

Beyond these classical SMBO paradigms, recent work explores zero- or few-shot HPO and the use of large language models (LLMs) as decision-makers within the HPO loop. The Pretrained Optimization Model (POM) attains compelling zero-shot performance across diverse benchmarks~\cite{li2024pom}. Zhang et al.~\cite{zhang2023llmhpo} further showed that LLMs can rival traditional HPO methods under limited search budgets. Pushing this direction, the SLLMBO framework~\cite{mahmammadli2024sllmbo} (\emph{Sequential Large Language Model-Based Hyperparameter Optimization}) couples an LLM policy with TPE to allocate budgets dynamically, outperforming standard BO baselines under constrained resources. These trends dovetail with broader AutoML efforts that seek end-to-end automation of the ML pipeline, where SMBO-style surrogates and LLM-driven priors increasingly coexist.

A popular line of HPO-related research refers to self-adaptation of hyperheuristic methods, e.g., Particle Swarm Optimization~\cite{Okulewiczetal2022,Okulewiczetal2020}. A separate stream of research encompasses portfolio-based hybridizations, where an ensemble of candidate algorithms is considered, with the aim of selecting an optimal parameterization of the algorithm that is best suited for the task considered~\cite{gecco2025,Zakrzewski2025KES,Zychowski2025KES,CHM,Zajecka20245775}.

Beyond hyperparameter optimization, AutoML frameworks aim to automate the entire machine learning pipeline, spanning data preprocessing, model selection, ensembling, and evaluation. For example, AutoML systems increasingly integrate multi-fidelity and early-stopping strategies to reduce computational overhead while maintaining competitive predictive accuracy~\cite{zoller2021benchmark}. More recent trends focus on incorporating foundation models and LLMs into AutoML workflows, enabling adaptive pipeline construction and even natural-language–driven configuration of experiments~\cite{guo2024uniautoml}. In addition, an efficiency-oriented AutoML system was introduced in~\cite{zhang2023etop}, leveraging surrogate modeling to minimize redundant evaluations. A Text-to-ML framework was presented in~\cite{xu2024large}, where LLMs generate full ML workflow code from textual descriptions and AutoML techniques subsequently evaluate and optimize these programs. These advances suggest a convergence of AutoML with general-purpose AI assistants, shifting the paradigm from purely search-based systems toward interactive, knowledge-driven automation.

The above trends align with a broader view of HPO that considers multiple objectives and advanced search strategies. In practical applications, hyperparameter tuning often involves trade-offs (e.g., predictive accuracy vs.\ model complexity or inference latency). Multi-objective optimization techniques have therefore been introduced into HPO to balance such competing criteria. For example, \emph{population-based} (swarm) approaches such as multi-objective PSO have been used to jointly optimize classification accuracy and model simplicity by treating feature selection as part of the hyperparameter search process~\cite{Xue2013PSO}. In parallel, advanced Bayesian optimization methods improve sample efficiency and robustness for design, tuning, and control tasks; e.g., the funneled BO variant shapes the search landscape to guide optimization more effectively in challenging settings~\cite{MartinezCantin2018}. By explicitly accounting for performance trade-offs or by enhancing search efficiency, these approaches help AutoML and HPO systems deliver models that are both accurate and resource-conscious.

\subsection{Tuning the Bootstrap Rate in Random Forests}

Despite the strong performance of RFs with default settings, studies have shown that careful hyperparameter tuning can yield additional performance gains~\cite{probst2019Hyperparameters}. In many cases, RF performs “reasonably well” with defaults (such as using all data in each bootstrap and $\sqrt{N}$ features per split), but fine-tuning can further improve metrics like prediction accuracy. A variety of strategies exist for RF hyperparameter optimization, including grid search, random search, and modern Bayesian optimization methods. For instance, the authors of~\cite{taiwo2025hyperparameter} applied grid search, randomized search, and TPE-based Bayesian optimization to tune RF hyperparameters for a human activity recognition task. This tuning achieved only a modest improvement in accuracy (~0.15\% increase, from 96.23\% to 96.37\%)---highlighting that RF defaults are often near-optimal---but came at the cost of greatly increased training time. Nonetheless, in other scenarios the impact of tuning is more pronounced. A recent survey on RF hyperparameters for fall-detection found optimal value ranges for each hyperparameter and noted that certain combinations can significantly boost performance compared to default settings (while other combinations can degrade it)~\cite{le2023survey}. This indicates that the benefit of tuning can be context-dependent: for some datasets and objectives, default parameters may suffice, whereas in others, carefully chosen hyperparameters make a substantial difference in predictive performance.

Researchers have also proposed enhanced RF variants to further improve performance. For example, an improved “broad granular random forest” algorithm was shown to outperform the standard RF on certain classification problems~\cite{xingyu2023bgrf}, and RFs have been used as high-quality surrogate models in optimization tasks~\cite{handing2020random}---both reflecting the continued interest in refining RF behavior. In general, however, such advances often build upon or benefit from judicious hyperparameter tuning of the standard RF.

While RFs have been extensively studied in the literature, the impact of the BR selection has received surprisingly little systematic attention. In the standard RF setting, each tree is trained on a bootstrap sample of the same size as the original dataset (i.e., BR~$=1$), a convention inherited from Breiman's original formulation~\cite{breiman2001random}. However, BR is a design choice parameter that can significantly affect the statistical properties and generalization ability of RF models.

Several studies have analyzed the role of subsampling, i.e., sampling without replacement with BR~$<$~1.0, as an alternative to bootstrapping. As shown in~\cite{duroux2016impact}, subsampling interacts with pruning in pure RFs, while the theoretical implications of subsampling versus bootstrapping in the context of infinite forests are examined in~\cite{scornet2016asymptotics}. Subsampling was also used to facilitate formal statistical inference: a new approach to variance estimation in RFs trained with subsampling was proposed in~\cite{wager2014confidence}, and in~\cite{mentch2016quantifying} it was demonstrated that U-statistics theory enables hypothesis testing and confidence interval construction for such models. However, these approaches focus exclusively on sampling without replacement and do not explore the full range of BR values in the bootstrapping setup.

In our recent paper~\cite{kazmierczak2024bootstrap}, we demonstrated that increasing the bootstrap sample size beyond the training set size (BR~$>1.0$) can improve performance in classification tasks. This finding challenges the common assumption that BR~$=1.0$ is near-optimal. While the study presents a formal theoretical argumentation for exploring BR $>1.0$, it nevertheless provides only a partial explanation of the results. Moreover, the focus on classification distinguishes that work from this regression-focused study.


The vast majority of studies using RFs for classification tasks do not explicitly tune the BR hyperparameter, leaving it at its default value of BR~$=$~1.0~\cite{Carreira-Perpinan_Hada_2023,Nguyen_Nguyen_Freedman_2024,Kasumba_Neumman_2024}. Examples of the few works that explore alternative BR values include~\cite{kazmierczak2024bootstrap} (BR $\in$ [0.2, 5.0]), \cite{martinez2010Out} (BR $\in$ [0.2, 1.2]), and \cite{Adnan2014improving} (BR $\in$ [0.78, 1.17]). We did not find any articles reporting BR values below 0.2. In the case of regression tasks, all the studies we found use the default BR value of 1.0, with representative examples including~\cite{fiedler2024using,Mao2024comparative,ali2024use,Venkatraman2024parsimonious}. Moreover, none of the popular ML libraries with RF implementations, such as scikit-learn~\cite{scikit-learn2011}, Weka~\cite{weka2009}, H2O.ai~\cite{h2o_package_or_module}, XGBoost~\cite{chen2016xgboost}, or LightGBM~\cite{ke2017light}, support bootstrap sample sizes larger than the training set size by default. This highlights that the space of BR~$>$~1.0 remains largely unexplored in practice.

\section{Experimental Setup}
\label{sec:ExperimentalSetup}
The experimental evaluation was conducted on 39 publicly available, popular regression datasets, selected to ensure diversity in the number of observations, feature types, and target domains. All datasets underwent a uniform preprocessing pipeline implemented in Python, employing pandas 2.2.2, NumPy 1.26.4, and scikit-learn 1.5.0. The preprocessing steps included:
\begin{itemize}
    \item Removal of duplicate rows;
    \item Elimination of columns with a single unique value;
    \item Exclusion of rows with missing target values;
    \item Imputation of missing numerical features using column means;
    \item Replacement of missing categorical features using a dedicated placeholder category;
    \item One-hot encoding of all categorical features: binary attributes were encoded using the drop-first strategy, while non-binary ones were fully expanded;
    \item All numerical features, as well as the target variable, were standardized using Z-score normalization (zero mean, unit variance).
\end{itemize}
Detailed characteristics of the datasets after preprocessing are provided in Table~\ref{tab:dataset_characteristics}.
\begin{table}
\centering
\caption{
Characteristics of the datasets. Subsequent columns indicate the dataset name, the number of numerical and binary features (including both original binary features and those resulting from one-hot encoding), and the number of observations, respectively. The datasets were collected from two sources: the UCI Machine Learning Repository (denoted by '(*)') and the Kaggle platform (the remaining ones). Kaggle datasets are hyperlinked, whereas those from UCI are not.}

\begin{tabular}{llll}
\toprule
Dataset & Num. Feat. & Bin. Feat. & Observations \\
\midrule
Abalone (*) & 7 & 3 & 4177 \\
\href{https://www.kaggle.com/datasets/ashydv/advertising-dataset}{Advertising} & 3 & 0 &  200 \\
Airfoil Self-Noise (*) & 5 & 0 &  1503 \\
Auction Verification (*) & 7 & 0 &  2043 \\
\href{https://www.kaggle.com/datasets/yasserh/auto-mpg-dataset}{Auto MPG} & 7 & 0 &  398 \\
\href{https://www.kaggle.com/datasets/dongeorge/beer-consumption-sao-paulo}{Beer Consumption} & 13 & 1 &  365 \\
\href{https://www.kaggle.com/datasets/ahmadrezagholami2001/bigmart-sales-dataset}{Bigmart Sales} & 3 & 4 &  4566 \\
\href{https://www.kaggle.com/datasets/fedesoriano/body-fat-prediction-dataset}{Body Fat} & 14 & 0 &  252 \\
\href{https://www.kaggle.com/datasets/fedesoriano/the-boston-houseprice-data}{Boston House Prices} & 13 & 0 &  506 \\
\href{https://www.kaggle.com/datasets/anubhabswain/brain-weight-in-humans}{Brain Weight} & 3 & 0 &  237 \\
\href{https://www.kaggle.com/datasets/jacopoferretti/child-weight-at-birth-and-gestation-details/data}{Child Weight at Birth} & 6 & 0 &  1236 \\
\href{https://www.kaggle.com/datasets/himelsarder/coffee-shop-daily-revenue-prediction-dataset}{Coffee Shop Revenue} & 6 & 0 &  2000 \\
Concrete Comp. Strength (*) & 8 & 0 &  1005 \\
\href{https://www.kaggle.com/datasets/japondo/corn-farming-data/data}{Corn Farming} & 6 & 29 &  422 \\
Daily Demand Forecasting (*) & 12 & 0 &  60 \\
Energy Efficiency (*) & 8 & 0 &  768 \\
Facebook Metrics (*) & 17 & 4 &  500 \\
\href{https://www.kaggle.com/datasets/vipullrathod/fish-market}{Fish Market} & 5 & 7 &  159 \\
\href{https://www.kaggle.com/datasets/denkuznetz/food-delivery-time-prediction}{Food Delivery Time} & 3 & 18 &  1000 \\
Forest Fires (*) & 12 & 0 & 513 \\
\href{https://www.kaggle.com/datasets/krupadharamshi/fuelconsumption}{Fuel Consumption} & 2 & 27 & 527 \\
Garment Productivity (*) & 14 & 7 & 1197 \\
\href{https://www.kaggle.com/datasets/rashadrmammadov/gun-price-prediction-linear-regression-model}{Gun Price} & 4 & 0 & 100 \\
\href{https://www.kaggle.com/datasets/sougatapramanick/happiness-index-2018-2019}{Happiness Index} & 6 & 0 & 312 \\
\href{https://www.kaggle.com/datasets/stealthtechnologies/regression-dataset-for-household-income-analysis}{Household Income} & 4 & 27 & 10000 \\
Infrared Thermography (*) & 30 & 15 & 1020 \\
\href{https://www.kaggle.com/datasets/gyanprakashkushwaha/laptop-price-prediction-cleaned-dataset}{Laptop Price} & 7 & 36 & 1272 \\
Liver Disorders (*) & 5 & 0 & 341 \\
\href{https://www.kaggle.com/datasets/harshsingh2209/medical-insurance-payout}{Medical Insurance} & 3 & 6 & 1337 \\
\href{https://www.kaggle.com/datasets/kukuroo3/mosquito-indicator-in-seoul-korea}{Mosquito Indicator Seoul} & 8 & 6 & 1295 \\
\href{https://www.kaggle.com/datasets/nehalbirla/motorcycle-dataset/data}{Motorcycle} & 3 & 5 & 1051 \\
\href{https://www.kaggle.com/datasets/sujithmandala/pokmon-combat-power-prediction}{Pok\'emon Combat Power} & 6 & 28 & 151 \\
\href{https://www.kaggle.com/datasets/abrambeyer/openintro-possum}{Possum Age} & 9 & 2 & 102 \\
Servo (*) & 2 & 10 & 167 \\
Student Performance (*) & 13 & 30 & 649 \\
\href{https://www.kaggle.com/datasets/denkuznetz/taxi-price-prediction}{Taxi Price} & 6 & 15 & 951 \\
\href{https://www.kaggle.com/datasets/ahmedaffan789/toy-cement-strength-data-set/data?select=TestData.csv}{Toy Cement Strength} & 8 & 0 & 809 \\
\href{https://www.kaggle.com/datasets/fedesoriano/wind-speed-prediction-dataset/data}{Wind Speed} & 11 & 6 & 6574 \\
Wine Quality (*) & 11 & 0 & 5318 \\
\bottomrule
\end{tabular}%
\label{tab:dataset_characteristics}
\end{table}

The performance of RF models depends on several key hyperparameters~\cite{Probst_2019}, including the number of trees in the ensemble (\texttt{nt}), the maximum depth of each tree (\texttt{md}), the minimum number of samples required to split an internal node (\texttt{mss}), the minimum number of samples required at a leaf node (\texttt{msl}), and the number of features considered when searching for the best split (\texttt{mf})\footnote{Hyperparameter notation follows \cite{kazmierczak2024bootstrap}.}. We adopted the default values from the scikit-learn 1.5.0 implementation of \texttt{RandomForestRegressor} as the baseline configuration, i.e., \texttt{nt = 100}, \texttt{md = None} (no maximum depth), \texttt{mss~=~2}, \texttt{msl = 1}, and \texttt{mf = 1.0} (100\% of features). We refer to this model as RF[100]. Based on this model, we constructed 15 alternative configurations, each differing from the baseline by exactly one hyperparameter value:
\begin{itemize}
    \item RF[200], RF[500]: number of trees increased to 200 and 500, respectively;
    \item RF[md10], RF[md15], RF[md20], RF[md25]: maximum tree depth limited to 10, 15, 20, and 25, respectively;
    \item RF[mss3], RF[mss4], RF[mss6], RF[mss8]: minimum samples required to split an internal node increased to 3, 4, 6, and 8, respectively;
    \item RF[msl2], RF[msl3], RF[msl4], RF[msl5]: minimum samples required at a leaf node increased to 2, 3, 4, and 5, respectively;
    \item RF[mfLog2]: number of features considered at each split changed to $\log_2(p)$.
\end{itemize}

Following~\cite{kazmierczak2024bootstrap}, we tested ten values of BR:
\[
\textrm{BR} \in \{0.2, 0.4, 0.6, 0.8, 1.0, 1.2, 2.0, 3.0, 4.0, 5.0\}.
\]
BR values less than or equal to 1.0 correspond to subsampling, while values greater than 1.0 result in oversampling of the training data. These settings affect both the size and composition of the bootstrap samples used during model fitting.
Each (RF model, BR) configuration pair was evaluated using 2-fold cross-validation repeated 50 times, yielding 100 evaluations per setup, measured by the mean squared error (MSE).


\section{Results}
\label{sec:Results}
For each dataset, we determined the combination of RF model configuration and BR value that yielded the lowest MSE. Table~\ref{tab:regression_results} summarizes these results.
\begin{table}
\centering
\caption{Regression results. Columns list the dataset name, the optimal RF configuration, the resulting MSE, the best BR value, and the p-value from the t-test, respectively.}
\begin{tabular}{lllll}
\toprule
Dataset & Best model & MSE & BR & p-value \\
\midrule
Abalone & RF[500] & 0.442 & 0.2 & $< 10^{-5}$ \\
Advertising & RF[100] & 0.069 & 1.0 & $0.14030$ \\
Airfoil Self-Noise & RF[500] & 0.091 & 2.0 & $< 10^{-5}$ \\
Auction Verification & RF[mss6] & 0.012 & 2.0 & $< 10^{-5}$ \\
Auto MPG & RF[500] & 0.137 & 0.8 & $0.03748$ \\
Beer Consumption & RF[500] & 0.352 & 0.4 & $< 10^{-5}$ \\
Bigmart Sales & RF[msl5] & 0.406 & 0.2 & $< 10^{-5}$ \\
Body Fat & RF[msl2] & 0.035 & 1.2 & $0.00217$ \\
Boston House Prices & RF[mfLog2] & 0.142 & 4.0 & $< 10^{-5}$ \\
Brain Weight & RF[mss6] & 0.406 & 0.2 & $< 10^{-5}$ \\
Child Weight at Birth & RF[msl3] & 0.752 & 0.2 & $< 10^{-5}$ \\
Coffee Shop Revenue & RF[200] & 0.056 & 0.6 & $< 10^{-5}$ \\
Concrete Comp. Strength & RF[mfLog2] & 0.116 & 3.0 & $< 10^{-5}$ \\
Corn Farming & RF[500] & 0.144 & 0.4 & $0.00119$ \\
Daily Demand Forecasting & RF[mfLog2] & 0.200 & 5.0 & $< 10^{-5}$ \\
Energy Efficiency & RF[100] & 0.003 & 1.2 & $0.19518$ \\
Facebook Metrics & RF[msl2] & 0.310 & 3.0 & $< 10^{-5}$ \\
Fish Market & RF[mfLog2] & 0.046 & 4.0 & $< 10^{-5}$ \\
Food Delivery Time & RF[500] & 0.273 & 0.4 & $< 10^{-5}$ \\
Forest Fires & RF[msl5] & 1.013 & 0.2 & $< 10^{-5}$ \\
Fuel Consumption & RF[500] & 0.158 & 1.0 & $0.59362$ \\
Garment Productivity & RF[500] & 0.526 & 0.6 & $0.00006$ \\
Gun Price & RF[msl2] & 0.204 & 0.4 & $0.00027$ \\
Happiness Index & RF[mfLog2] & 0.192 & 2.0 & $0.00602$ \\
Household Income & RF[msl5] & 0.953 & 1.0 & $0.00372$ \\
Infrared Thermography & RF[500] & 0.246 & 0.6 & $< 10^{-5}$ \\
Laptop Price & RF[mfLog2] & 0.132 & 1.2 & $0.00003$ \\
Liver Disorders & RF[msl4] & 0.831 & 0.2 & $< 10^{-5}$ \\
Medical Insurance & RF[msl3] & 0.142 & 0.2 & $< 10^{-5}$ \\
Mosquito Indicator in Seoul & RF[200] & 0.091 & 1.2 & $0.00007$ \\
Motorcycle & RF[mss8] & 0.358 & 0.8 & $0.02916$ \\
Pok\'emon Combat Power & RF[mfLog2] & 0.143 & 4.0 & $0.00689$ \\
Possum Age & RF[msl5] & 0.818 & 0.6 & $< 10^{-5}$ \\
Servo & RF[100] & 0.176 & 1.2 & $0.03034$ \\
Student Performance & RF[500] & 0.707 & 0.2 & $< 10^{-5}$ \\
Taxi Price & RF[200] & 0.070 & 0.8 & $0.08948$ \\
Toy Cement Strength & RF[mfLog2] & 0.133 & 3.0 & $0.00003$ \\
Wind Speed & RF[500] & 0.748 & 0.4 & $0.40784$ \\
Wine Quality & RF[mfLog2] & 0.627 & 1.0 & $0.96039$ \\
\bottomrule
\end{tabular}%
\label{tab:regression_results}
\end{table}

\subsection{Performance Ranking of RF Configurations}
Among the 16 RF configurations evaluated, only a subset consistently achieved top results across multiple datasets. RF[500] emerged as the most dominant, securing first place for 11 datasets, followed by RF[mfLog2] with 9 wins. RF[msl5] led on 4 datasets, while RF[100], RF[200], and RF[msl2] each topped 3 datasets. RF[msl3] and RF[mss6] performed best on 2 datasets each, and each of RF[msl4] and RF[mss8] achieved a single dataset win. Subsequent analysis focuses on the leading configurations, defined as those with top performance on at least 2 datasets---altogether, there were 8 such configurations.

\subsection{Impact of BR~\texorpdfstring{$>$}{>}~1.0 vs. BR~\texorpdfstring{$\leq$}{<=}~1.0 on RF Performance}
The best-performing setup used BR~$>$~1.0 in 15 out of 39 datasets, while BR~$\leq$~1.0 yielded the lowest MSE for 24 datasets. To assess whether these results are statistically significant, we performed one-sided paired t-tests. For each dataset, we compared the best-performing (lowest MSE) configuration from one BR group against all configurations from the opposite group, testing whether the winning group achieved significantly lower MSE. More precisely, we tested whether the mean MSE value over 100 runs for the best-performing configuration was statistically lower than the mean MSE values of all configurations with a BR value from the opposite group.

The last column of Table~\ref{tab:regression_results} reports the maximum p-value from the aforementioned comparisons for each dataset. To assess statistical strength, we considered six significance thresholds: 0.1, 0.05, 0.01, 0.001, 0.0001, and 0.00001. For each threshold, we counted only the datasets in which the performance difference was statistically significant (i.e., the p-value was below the threshold). At the 0.1 level, BR~$\leq$~1.0 yielded significantly better results for 20 datasets, while BR~$>$~1.0 was superior for 14 ones, resulting in a net difference of $-6$. This pattern persisted across stricter thresholds, with net differences of -5, -4, -5, -4, and -6 at the 0.05, 0.01, 0.001, 0.0001, and 0.00001 levels, respectively.

In 38.5\% of the analyzed datasets, the BR hyperparameter in the best-performing configuration had a value greater than 1.0. Moreover, when considering only statistically significant results, BR~$>$~1.0 accounted for between 40.0\% and 43.3\% of the top configurations, depending on the chosen p-value threshold. These findings clearly indicate that, contrary to common belief, it is worth exploring BR~$>$~1.0 during the hyperparameter optimization of RFs.

\subsection{Patterns in BR Curves}
\begin{figure*}
\centering
\includegraphics[width=0.9999\textwidth]{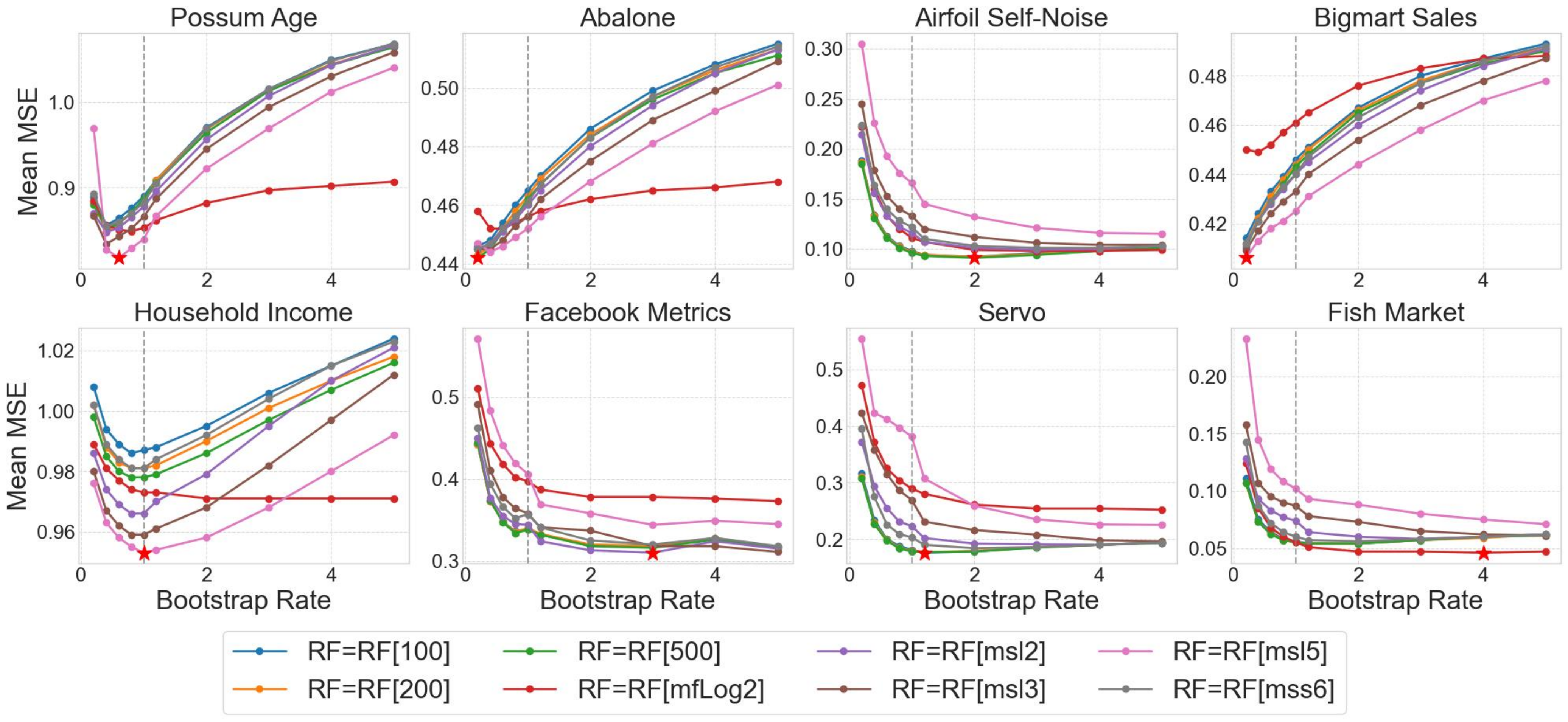}
\caption{BR curve profiles for selected representative datasets. For each dataset, the optimal configuration is marked with a red star. The rest of the plots for other datasets are provided in \hyperref[sec:br_curves]{Appendix}.}
\label{fig:BR_curves}
\end{figure*}

Fig.~\ref{fig:BR_curves} depicts the relationship between the BR and the mean MSE for selected datasets and the top-performing models. In what follows, we refer to these curves as BR curves. As expected, the BR curves exhibit distinct shapes for different datasets. Nevertheless, three general patterns can be distinguished. The first pattern, exemplified by the Bigmart Sales dataset, suggests that BR values below 0.2, i.e., lower than those tested here, might yield even better performance. The second pattern, observed, for example, in Household Income and Possum Age, is characterized by a decrease in MSE starting from BR~$=$~0.2, reaching an optimum before BR~$=$~1.0, and increasing thereafter. The third pattern occurs when the BR range between 0.2 and 1.0 is insufficient to reach the optimum; for instance, this can be seen in the Fish Market dataset.

The above findings emphasize the importance of tailoring the BR parameter to the specific dataset rather than relying on package defaults (where commonly BR~$=$~1.0). A partial explanation of the observed differences in BR curve shapes and optimal BR values is presented in Section~\ref{syntetic_data_test}.
\subsection{Optimal BR Distribution Across RF Configurations}
\begin{figure}
\centering
\includegraphics[width=0.489\textwidth]{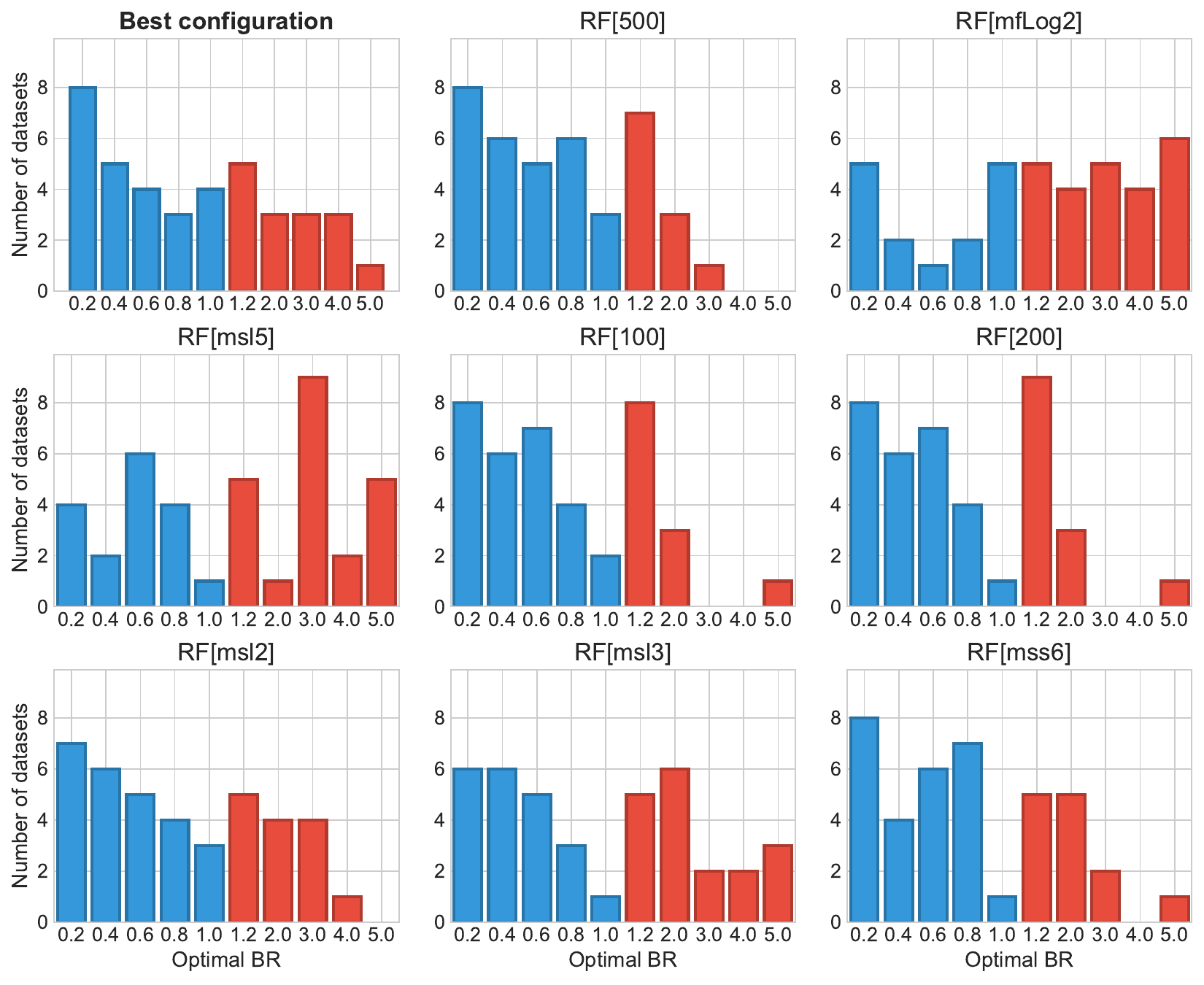}
\caption{Distribution of optimal BR values in the best configuration (top left) and across individual RF setups.}
\label{fig:model_br}
\end{figure}

Fig.~\ref{fig:model_br} shows the frequency of BR wins, both overall and under selected RF hyperparameter configurations. The first key observation is that the distribution of winning BR values varies significantly with the RF configuration. As the minimum number of observations required to form a leaf increases, configurations with BR~$>$~1.0 more frequently become optimal. For RF[msl2], RF[msl3], and RF[msl5], the number of datasets for which configurations with BR~$>$~1.0 achieved the best performance equals 14, 18, and 22, respectively. Apparently, for a substantial portion of the datasets, a low number of training instances combined with the constraint on the minimum number of observations per leaf led to overly simplistic and underfitted trees. In such cases, a higher BR effectively increased the number of observations used to train each tree, allowing it to be more complex, i.e., deeper and with more leaves.

An even stronger preference for higher BR values is exhibited by RF[mfLog2], for which configurations with BR~$>$~1.0 yielded the best results for as many as 24 datasets. Generally, the effectiveness of the ensemble of ML models is primarily determined by two factors: diversity and the strength of individual base models. The small size of feature subsets considered for splits in the RF[mfLog2] configuration promotes high diversity but simultaneously reduces the strength of base learners---since node splits are based on fewer candidate features, their ability to minimize MSE is impaired compared to configurations with larger candidate subsets. We hypothesize that RF[mfLog2] favors higher BR values in order to increase the strength of base models. This is in line with the general principle that the strength of an individual model increases with the number of distinct training observations it sees.

Interestingly, BR~$=$~1.0, i.e., the original bootstrap formulation, most commonly used in the literature and set as the default in most ML libraries, yielded the best performance for only 4 out of 39 datasets. When averaged across all evaluated configurations, BR~$=$~1.0 was optimal for just 2.2 datasets (5.6\% of the total). Notably, BR~$=$~1.2, a nearby value, achieved more wins than BR~$=$~1.0 both globally and across all configurations, with the exception of RF[mfLog2], where the number of wins was identical.

Finally, BR~$=$~0.2, i.e., the smallest value considered in our study, was the most frequently optimal configuration, suggesting that in many cases BR values below 0.2 may yield even better results. Consequently, when tuning RFs, it is advisable to explore not only BR~$>$~1.0, which is one of the main takeaways of this work, but also BR~$<$~0.2, which remain largely underexplored in the literature.

Consistent with our previous work~\cite{kazmierczak2024bootstrap}, this study further confirms that selecting BR values greater than 1.0 can yield superior performance also in the regression setting, although the gains observed here are slightly smaller than those reported for classification tasks.

\section{Discussion}
\label{sec:Discussion}
\subsection{The impact of data characteristics on the BR selection}  
To better understand when and why certain BR values perform better, we examined how dataset-specific characteristics correlate with the preference for low or high BRs. For each dataset, we computed a wide range of structural, statistical, and information-theoretic metrics and compared their distributions between datasets that favor lower BRs ($\leq$~1.0) vs. higher BRs ($>$~1.0). We applied Mann--Whitney~$U$ tests for statistical significance, computed Cohen's $d$ for effect sizes, and analyzed Spearman correlations with optimal BR groups. While the Mann--Whitney~$U$ test highlights statistically significant differences between datasets grouped by optimal BR levels, Spearman’s rank correlation captures monotonic trends in metric values across the BR spectrum. Together, these methods offer complementary insights into how dataset characteristics relate to BR preferences.

\begin{table*}
\centering
\caption{Comparison of selected dataset-level and model-level characteristics between groups favoring low ($\leq 1.0$) and high ($> 1.0$) BR. The columns show the $p$-values from Mann--Whitney~$U$ tests, Cohen’s $d$ effect sizes, Spearman correlation coefficients ($\rho$), and the associated $p$-values.}
\label{tab:br_metrics}
\begin{tabular}{llrrrr}
\toprule
 & Metric & MW $p$-value & Cohen’s $d$ & Spearman $\rho$ & Spearman $p$-value \\
\midrule
\multirow{6}{*}{\textbf{Dataset-level}}
 & Mutual information (sum) & 0.004 & 0.89 & +0.468 & 0.0026 \\
 & Mean target variance in kNN & 0.009 & -0.97 & -0.426 & 0.0068 \\
 & HSIC (linear kernel) & 0.019 & 0.41 & +0.384 & 0.0158 \\
 & HSIC (RBF kernel) & 0.036 & 0.36 & +0.342 & 0.0332 \\
 & HSIC (Laplace kernel) & 0.051 & 0.37 & +0.318 & 0.0482 \\
 & High feature--target correlation & 0.118 & 0.53 & +0.257 & 0.1136 \\
\midrule
\multirow{4}{*}{\textbf{Model-level}}
 & OOB $R^2$ & $< 10^{-5}$ & 1.95 & +0.838 & $< 10^{-5}$ \\
 & Full-model $R^2$ & 0.00012 & 1.47 & +0.628 & 0.00002 \\
 & Adjusted $R^2$ & 0.00015 & 1.44 & +0.618 & 0.00003 \\
 & Signal-to-noise ratio & 0.00001 & 0.55 & +0.716 & $< 10^{-5}$ \\
\bottomrule
\end{tabular}
\end{table*}

The most informative metrics are summarized in Table~\ref{tab:br_metrics}. The \textit{sum of mutual information} exhibited the strongest positive correlation with optimal BR (Spearman $\rho = 0.468$, $p = 0.0026$) and showed a large effect size ($d = 0.89$), suggesting that datasets with higher overall feature–target dependence tend to favor higher BR values. Conversely, the \textit{mean target variance in kNN (k=10)}, which captures local target variability, showed a strong negative correlation ($\rho = -0.426$, $p = 0.0068$), indicating that datasets with more locally diverse targets benefit from lower BRs, likely because using less resampled data reduces the risk of overfitting.

Other dependence measures such as \textit{HSIC} (linear, RBF, and Laplace kernels) and \textit{High feature--target correlation} (defined as the number of columns in the dataset whose absolute correlation with the target variable exceeds 0.9) also showed moderate positive associations with BR, further supporting the idea that strong global signal structures make models more resilient to aggressive (BR~$>$~1.0) resampling. These findings highlight that the optimal BR depends on the intrinsic properties of the dataset, with low BRs being more effective at capturing local data variation and high BRs better exploiting strong global dependencies between features and targets.

\subsection{Model-level metrics and the BR selection}
In addition to intrinsic dataset characteristics, we investigated the relationship between model-level indicators and the choice of optimal BR. Specifically, we analyzed metrics reflecting model performance and robustness, including out-of-bag $R^2$, full-model $R^2$, adjusted $R^2$, and the signal-to-noise ratio (SNR). As shown in Table~\ref{tab:br_metrics}, all of these metrics exhibited strong positive Spearman correlations with the optimal BR value ($\rho \ge 0.618$, $p \le 0.00003$ for all), suggesting that models which already generalize well or capture strong predictive signals tend to benefit from higher BRs. This pattern is further supported by large effect sizes (Cohen's $d$ up to $1.95$) and significant group differences based on the Mann--Whitney~$U$ test. These findings suggest that models with high explanatory power and robustness (e.g., high $R^2$, high out-of-bag performance) can tolerate or benefit from heavier resampling.

\subsection{Tests on synthetic data}
\label{syntetic_data_test}
Based on the metrics obtained, we decided to investigate the behavior of the optimal BR$_{OPT}$ on synthetic datasets corrupted by varying levels of noise. In our experiments, datasets 
with BR$_{OPT} > 1.0$ yielded a mean coefficient of determination of $R^2 = 0.909$, whereas those with 
BR$_{OPT} < 1.0$ had a mean $R^2 = 0.545$. Hence, one may hypothesize that
\[
\text{BR}_{OPT} > 1.0 
\;\Longrightarrow\;
R^2 \approx 1
\;\Longrightarrow\;
\text{Low–noise data}.
\]
The converse implication does not necessarily hold.

\paragraph{Definition of Regression Tree} 
Given a training set  
\begin{equation}
\mathcal{D} = \{(x_i, y_i)\}_{i=1}^n,\quad x_i\in\mathbb{R}^p,\;y_i\in\mathbb{R},
\end{equation}
we seek a model of the form  
\begin{equation}
\hat f(x) = \sum_{m=1}^M c_m\,\mathbf{1}\{x \in R_m\},
\end{equation}
where \(\{R_1,\dots,R_M\}\) is a partition of \(\mathbb{R}^p\) into \(M\) disjoint axis-aligned hyperrectangles (each corresponding to a terminal leaf of the tree), and each \(c_m\) is the constant prediction on region \(R_m\). The constant \(c_m\) is given by:  
\begin{equation}
c_m = \frac{1}{|\mathcal{D}_m|}\sum_{x_i\in R_m}y_i,
\end{equation}
where \(|\mathcal{D}_m|\) is the number of training points in \(R_m\).

\paragraph{Synthetic data generation}
To evaluate the behavior of RFs under different noise levels, we generated synthetic data by partitioning the feature space into hyperrectangles defined by both categorical and numerical features. Specifically, we considered a feature space comprising two binary categorical features, denoted \(\mathrm{cat}_1\) and \(\mathrm{cat}_2\), each taking values in \(\{0, 1\}\), and two numerical features, \(\mathrm{num}_1 \in [0, 9]\) and \(\mathrm{num}_2 \in [0, 10]\). The numerical features were partitioned into intervals: \(\mathrm{num}_1\) was divided into three intervals \([0, 3]\), \([3, 6]\), and \([6, 9]\), while \(\mathrm{num}_2\) was divided into two intervals \([0, 5]\) and \([5, 10]\). This partitioning, combined with all possible combinations of the binary categorical features, resulted in \(2 \times 2 \times 3 \times 2 = 24\) distinct regions. Each region corresponds to a unique hyperrectangle in the feature space, collectively forming a partition of the space. For example, one such region \(R\) might be specified as
\begin{align*}
&R = \{(\mathrm{cat}_1, \mathrm{cat}_2, \mathrm{num}_1, \mathrm{num}_2) \mid \\ &\mathrm{cat}_1 = 0, \mathrm{cat}_2 = 1, \mathrm{num}_1 \in [0, 3], \mathrm{num}_2 \in [5, 10] \}.
\end{align*}
For each region \(R_m\), \(m = 1, 2, \dots, 24\), we generated 15 data points. Each data point \(x_i^{(m)} = (\mathrm{cat}_1, \mathrm{cat}_2, \mathrm{num}_1, \mathrm{num}_2)\) in region \(R_m\) was created by setting the categorical features to the values defining the region, i.e., \(\mathrm{cat}_1 = a_m\) and \(\mathrm{cat}_2 = b_m\) where \(a_m, b_m \in \{0, 1\}\), and sampling the numerical features uniformly from their respective intervals, i.e., \(\mathrm{num}_1 \sim \mathcal{U}(I_{1,m})\) and \(\mathrm{num}_2 \sim \mathcal{U}(I_{2,m})\), where \(I_{1,m}\) and \(I_{2,m}\) denote the intervals for \(\mathrm{num}_1\) and \(\mathrm{num}_2\) in region \(R_m\), respectively. For each region \(R_m\), a constant \(c_m\), representing the expected prediction for that region, was independently drawn from a uniform distribution over \([0, 10]\), i.e., \(c_m \sim \mathcal{U}(0, 10)\). Then, for each data point \(x_i^{(m)} \in R_m\), the corresponding response \(y_i^{(m)}\) was generated as $y_i^{(m)} = c_m + \varepsilon_i^{(m)}$, where \(\varepsilon_i^{(m)} \sim \mathcal{N}(0, \sigma^2)\) represents Gaussian noise with variance \(\sigma^2\). The noise terms \(\varepsilon_i^{(m)}\) are independent across all data points. In total, the synthetic dataset consists of \(24 \times 15 = 360\) data points, each comprising a feature vector \(x_i\) and a response \(y_i\), generated according to the procedure outlined above.

\begin{figure}[t]
    \centering
    \includegraphics[width=1\linewidth]{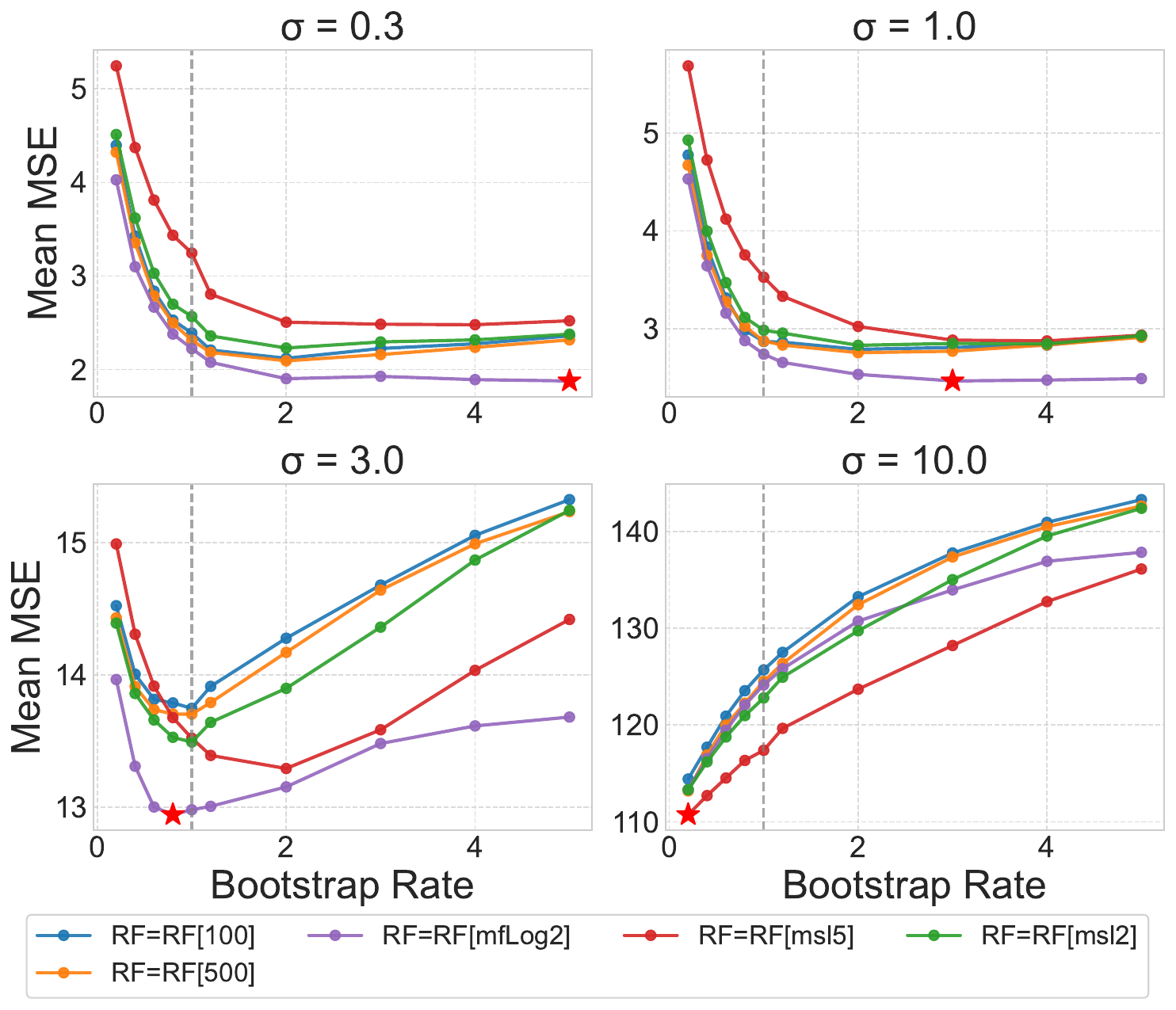}
    \caption{Test MSE of RFs trained on synthetic datasets under varying noise levels (\(\sigma\)) and BRs. Each curve represents a different noise setting. The results show that higher BRs perform better when noise is low, while lower BRs are better under high noise. This illustrates the trade-off between resampling intensity and noise sensitivity in model performance.}
    \label{fig:synthetic_results}
\end{figure}

\paragraph{Results}
We experimented with various values of \(\sigma\) to observe how RFs behave under different noise levels and to identify optimal BRs. Fig.~\ref{fig:synthetic_results} shows BR curves of RFs trained on the synthetic datasets.

\paragraph{Explanation of the bias–variance trade-off under varying bootstrap rate}
Let \(Y\) be the target value:
\begin{equation}
   Y \;=\; f(X)+\varepsilon,
   \qquad
   E[\varepsilon]=0,
   \qquad
   Var(\varepsilon)=\sigma^{2}
\end{equation}
and let $\widehat f_{b,T}(x)$ denote the RF predictor obtained by averaging the outputs of $T$ trees, each trained on a bootstrap sample containing $bn$ observations, where $b$ is the BR and $n$ is the training set size:
\begin{equation}
   \widehat f_{b,T}(x)\;=\; \frac{1}{T}\sum_{t=1}^{T}\widehat f_{b}^{(t)}(x).
\end{equation}
For a fixed feature vector $x$, the test mean-squared error decomposes into:
\begin{equation}
\begin{split}
\mathrm{MSE}_{b,T}(x)
   &= 
   \underbrace{\bigl(E[\widehat f_{b,T}(x)] - f(x)\bigr)^{2}}_{\text{Bias$^{2}$}} \\
   &+ 
   \underbrace{Var\!\bigl(\widehat f_{b,T}(x)\bigr)}_{\text{Aggregation variance}} 
   + \sigma^{2}.
\end{split}
\label{eq:bias_variance}
\end{equation}

\noindent Define
\begin{align}
\sigma^{2}_{\mathrm{tree}}(b,x) &:= Var\!\bigl(\widehat f_{b}^{(t)}(x)\bigr),\\
\rho(b,x) &:= Corr\!\bigl(\widehat f_{b}^{(t)}(x),\,\widehat f_{b}^{(s)}(x)\bigr), \quad t \neq s.
\end{align}
Then, for an ensemble of $T$ trees, the variance of the aggregated prediction can be expressed using the classical formula for the variance of an average of correlated random variables~\cite{hastie2009elements}:
\begin{equation}
\begin{aligned}
Var\!\bigl(\widehat f_{b,T}(x)\bigr)
&= \frac{\sigma^{2}_{\mathrm{tree}}(b,x)}{T}
  + \Bigl(1-\frac{1}{T}\Bigr)\rho(b,x)\,\sigma^{2}_{\mathrm{tree}}(b,x)\\
&\xrightarrow[T \to \infty]{}\;
\rho(b,x)\,\sigma^{2}_{\mathrm{tree}}(b,x).
\end{aligned}
\end{equation}

In our synthetic-data study (Fig.~\ref{fig:synthetic_results}), we observed that when the observation noise level $\sigma^2$ is small, trees have little noise to fit and $\sigma^2_{\mathrm{tree}}(b,x)$ remains modest. Consequently, the product $\rho(b,x)\,\sigma^2_{\mathrm{tree}}(b,x)$ appears to grow only mildly with $b$, while the bias can decrease noticeably, so increasing $b$ tends to reduce the overall MSE. Conversely, under higher noise, large $b$ can encourage fitting noise, making $\rho(b,x)\,\sigma^{2}_{\mathrm{tree}}(b,x)$ grow quickly. Using smaller $b$ increases diversity among trees, which helps reduce $Var(\widehat{f}_{b,T}(x))$ at the cost of higher bias, yet in our setup this trade-off still lowered the forest’s MSE.

\paragraph{Illustrative example}
To ground the above theoretical reasoning, consider the extreme case of pure noise and a single predictor \(x\) with responses $Y$.  
\begin{equation}
Y\sim\mathcal{N}(\mu,\sigma^2).
\end{equation}
The MSE for a prediction \(\hat y\) equals  
\begin{equation}
E\bigl[Y-\hat y\bigr]^2
= (E[Y]-\hat y)^2 + Var(Y)
= (\mu-\hat y)^2 + \sigma^2,
\end{equation}
which is minimized by \(\hat y=\mu\), giving \(\mathrm{MSE}=\sigma^2\).
Hence, the optimal predictor is \(\hat y(x)=\mu\), with zero variance in the predictor itself.

Suppose that we predict at a new point \(x_{\mathrm{test}}\), where
\begin{equation}
Y_{\mathrm{test}}\sim\mathcal{N}(\mu,\sigma^2).
\end{equation}
Consider trees that are deep-grown with \texttt{msl}=1. When BR is small, individual trees give various noisy predictions \(\widehat Y^{(t)}\)  for the same \(x_{\mathrm{test}}\), but the forest aggregates them via  
\begin{equation}
\hat Y_{\mathrm{RF}}=\frac{1}{T}\sum_{t=1}^T\widehat Y^{(t)}\approx E[Y]=\mu,
\end{equation}
which is the best possible prediction.

In the single-predictor scenario with \texttt{msl}=1, for a test point \(x_{\mathrm{test}}\), a regression tree predicts the $y$ value of the training point within that tree's bootstrap sample that is closest to \(x_{\mathrm{test}}\) in Euclidean distance.
Let $x_{\text{NN}}$ denote the global nearest neighbor of \(x_{\mathrm{test}}\) in the full training set,
with target $y_{\text{NN}}$. If a tree's bootstrap sample contains $x_{\text{NN}}$, then
$x_{\text{NN}}$ is also the closest point within that sample, and the tree predicts $y_{\text{NN}}$. For example, with $\mathrm{BR}=10$ and $T=100$ trees, the probability that $x_{\text{NN}}$ is included in every tree is $\bigl(1-e^{-10}\bigr)^{100}\approx 0.995$ for large $N$ (Eq.~\eqref{eq:limit}). Therefore, for high BR, it is likely that every tree (or at least most of them) will predict the same value at $x_{\mathrm{test}}$, namely $y_{\mathrm{NN}}$. In such a case \(\hat y\), which is the mean of all predictions, is approximately equal to the prediction of a single tree. This single-tree prediction is itself a random variable \(\widehat{Y}\) (used in training dataset), distributed as:
\begin{equation}
\widehat{Y}\sim\mathcal{N}(\mu,\sigma^2),
\end{equation}
and is independent of the new observation
\begin{equation}
Y_{\mathrm{test}}\sim\mathcal{N}(\mu,\sigma^2).
\end{equation}
Therefore,
\begin{equation}
\begin{aligned}
E(Y_{\mathrm{test}}-\hat Y)^2 &= (E[Y_{\mathrm{test}}-\hat Y])^2 + Var(Y_{\mathrm{test}}-\hat Y) \\ &= (\mu-\mu)^2 + 2\sigma^2 = 2\sigma^2.
\end{aligned}
\end{equation}

\begin{figure}
    \centering
    \includegraphics[width=1\linewidth]{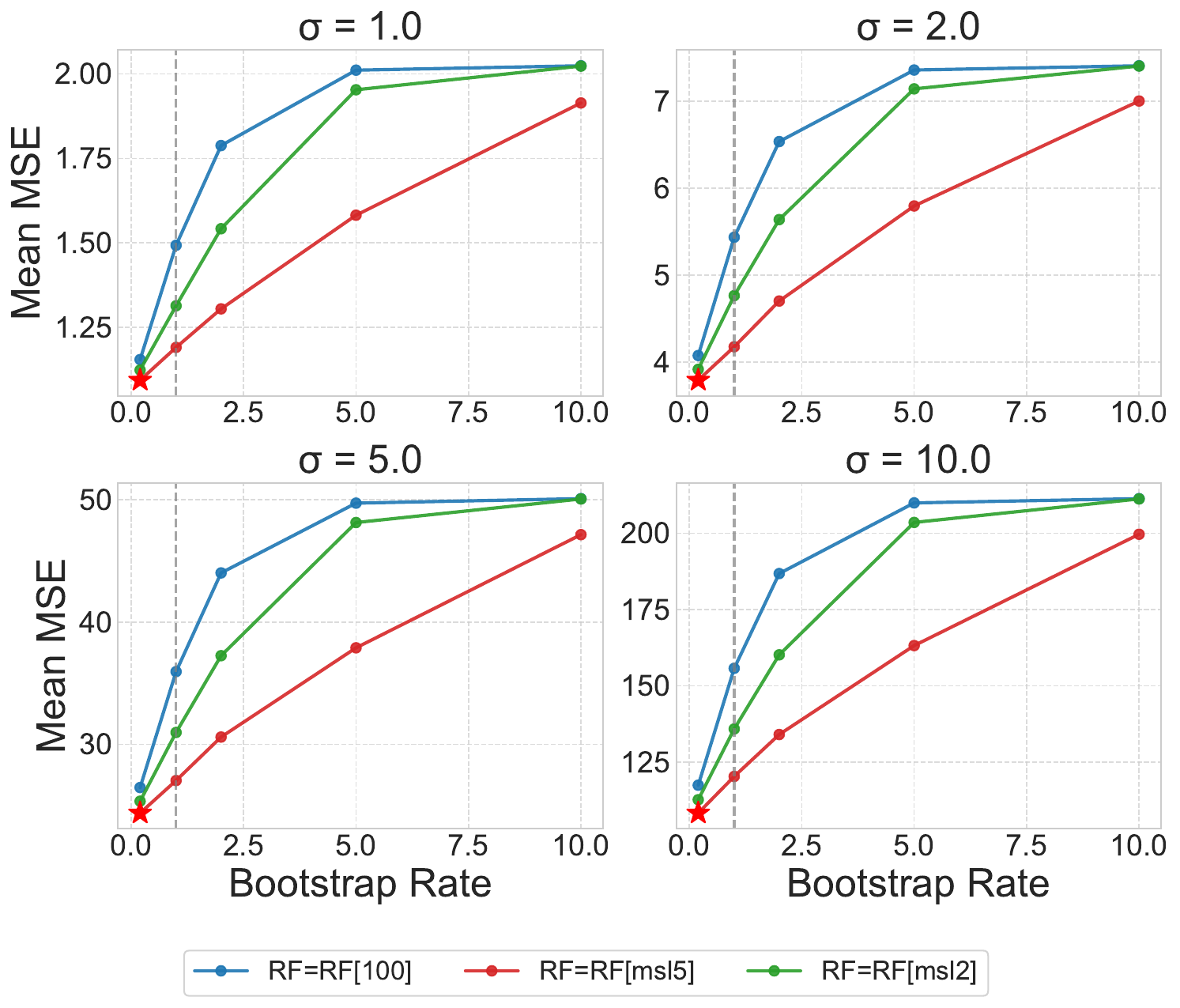}
    \caption{MSE across various BRs \(b\) and noise levels \(\sigma\) in pure-noise scenario.}
    \label{fig:noise_results}
\end{figure}

To verify these expectations, we carried out experiments for four noise levels $\sigma \in \{1.0, 2.0, 5.0, 10.0\}$. For each chosen $\sigma$ we generated $n = 1000$ i.i.d.\ samples $(x,y)$ with $x \sim \mathcal{U}[0,5]$ and $y \sim \mathcal{N}(0,\sigma^{2})$. The empirical results match the above theoretical considerations. As shown in Fig.~\ref{fig:noise_results}, the best-performing configuration employed a BR of 0.2 across all noise levels, attaining mean-squared errors (MSE) of 1.09 ($\sigma=1$), 3.79 ($\sigma=2$), 24.4 ($\sigma=5$), and 108 ($\sigma=10$). Conversely, the poorest configuration, using a BR of 10, produced MSE values of 2.02 ($\sigma=1$), 7.41 ($\sigma=2$), 50.1 ($\sigma=5$), and 211 ($\sigma=10$). These findings confirm that large BRs give $\text{MSE} \approx 2\sigma^{2}$, whereas small BRs achieve $\text{MSE} \approx \sigma^{2}$ in this pure-noise setting.

\section{Conclusions and Future Work}
\label{sec:Conclusions}
This study presents the first large-scale assessment of how the \emph{bootstrap rate} shapes RF regression performance. The main conclusion is that the long-standing default of \( \mathrm{BR}=1.0\) is not universally optimal. In 24 out of 39 real-world regression datasets the minimum mean-squared error (MSE) is obtained with \( \mathrm{BR}\le 1.0 \). In the remaining 15 datasets, i.e., more than one-third of the cases, the lowest MSE occurs at \(\mathrm{BR}>1.0\). Paired t-tests showed statistically significant differences between the \(\mathrm{BR}\le 1.0\) and \(\mathrm{BR}>1.0\) scores for most of the tested datasets, underscoring that careful BR tuning can yield non-trivial performance gains.

Strong global feature–target dependence (e.g., high cumulative mutual information or HSIC) correlates with a preference for larger BRs, because the extra data per tree reduces bias without being overwhelmed by noise. High local target variance favors smaller BRs, where additional tree diversity combats overfitting. Likewise, models that already exhibit high explanatory power (e.g., high out-of-bag \(R^2\)) tend to benefit from larger BRs.
        
Fewer samples per tree primarily lower variance through ensemble diversification; more samples per tree primarily lower bias by letting each tree learn richer structure. The optimal BR to some extent reflects the dataset’s signal-to-noise profile.

In future work, we plan to verify whether the observed relationship between optimal BR and data noise is present in the results of ensemble models other than RFs. Such an observation could signify the utility of BR tuning as a general ensemble principle. We also recommend that developers of ML and AutoML libraries incorporate support for tuning the BR, including values greater than 1.0, to enable users to fully leverage its impact on RF performance.

\bibliographystyle{IEEEtran}
\bibliography{references}

\onecolumn

\section*{APPENDIX: BR Curves}
\label{sec:br_curves}
\begin{figure}[H]
\centering
\includegraphics[width=1\textwidth]{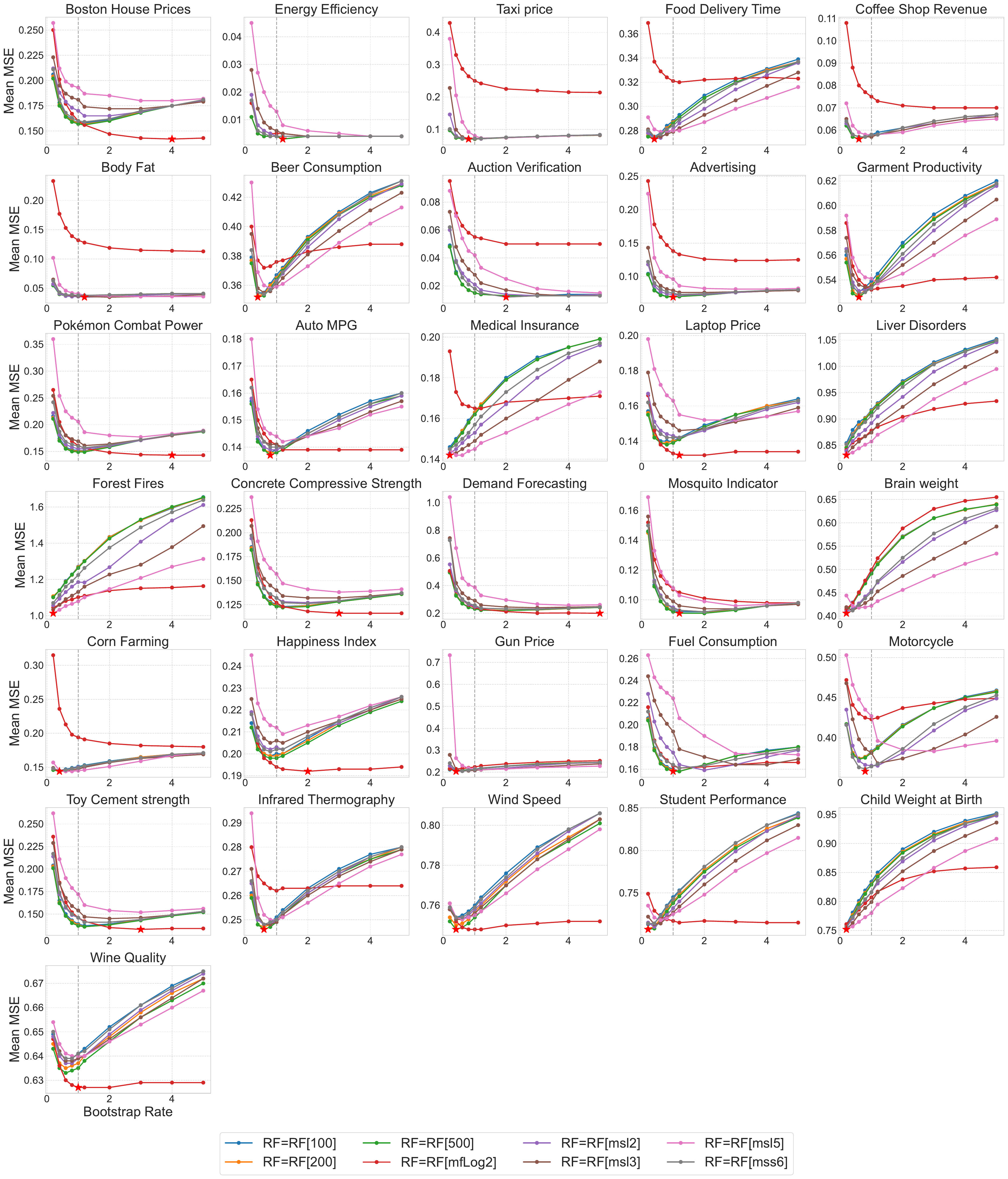}
\caption{Additional BR curves not included in Fig.~\ref{fig:BR_curves}.}
\end{figure}

\clearpage
\section*{APPENDIX: Detailed Results}

\begin{table}[h]
\addtolength{\tabcolsep}{-4pt}
\centering
\caption{Classification accuracy (mean $\pm$ std) for the Possum Age dataset (Part 1).}
\begin{tabular}{c|ccccccccc}
 \toprule
\diagbox{BR}{Model} & RF[100] & RF[200] & RF[500] & RF[md10] & RF[md15] & RF[md20] & RF[md25] & RF[mss3] & RF[mss4] \\
 \midrule
0.2 & 0.889 $\pm$ 0.108 & 0.882 $\pm$ 0.108 & 0.880 $\pm$ 0.103 & 0.889 $\pm$ 0.108 & 0.889 $\pm$ 0.108 & 0.889 $\pm$ 0.108 & 0.889 $\pm$ 0.108 & 0.889 $\pm$ 0.108 & 0.889 $\pm$ 0.107 \\
0.4 & 0.856 $\pm$ 0.102 & 0.849 $\pm$ 0.103 & 0.852 $\pm$ 0.103 & 0.856 $\pm$ 0.102 & 0.856 $\pm$ 0.102 & 0.856 $\pm$ 0.102 & 0.856 $\pm$ 0.102 & 0.855 $\pm$ 0.101 & 0.856 $\pm$ 0.103 \\
0.6 & 0.864 $\pm$ 0.106 & 0.858 $\pm$ 0.105 & 0.858 $\pm$ 0.105 & 0.864 $\pm$ 0.106 & 0.864 $\pm$ 0.106 & 0.864 $\pm$ 0.106 & 0.864 $\pm$ 0.106 & 0.864 $\pm$ 0.109 & 0.863 $\pm$ 0.109 \\
0.8 & 0.876 $\pm$ 0.113 & 0.870 $\pm$ 0.108 & 0.871 $\pm$ 0.108 & 0.876 $\pm$ 0.113 & 0.876 $\pm$ 0.113 & 0.876 $\pm$ 0.113 & 0.876 $\pm$ 0.113 & 0.874 $\pm$ 0.112 & 0.874 $\pm$ 0.113 \\
1 & 0.890 $\pm$ 0.116 & 0.885 $\pm$ 0.115 & 0.887 $\pm$ 0.113 & 0.890 $\pm$ 0.116 & 0.890 $\pm$ 0.116 & 0.890 $\pm$ 0.116 & 0.890 $\pm$ 0.116 & 0.890 $\pm$ 0.114 & 0.889 $\pm$ 0.116 \\
1.2 & 0.909 $\pm$ 0.120 & 0.909 $\pm$ 0.117 & 0.906 $\pm$ 0.116 & 0.909 $\pm$ 0.120 & 0.909 $\pm$ 0.120 & 0.909 $\pm$ 0.120 & 0.909 $\pm$ 0.120 & 0.908 $\pm$ 0.120 & 0.907 $\pm$ 0.119 \\
2 & 0.970 $\pm$ 0.134 & 0.968 $\pm$ 0.134 & 0.964 $\pm$ 0.136 & 0.970 $\pm$ 0.134 & 0.970 $\pm$ 0.134 & 0.970 $\pm$ 0.134 & 0.970 $\pm$ 0.134 & 0.969 $\pm$ 0.137 & 0.971 $\pm$ 0.137 \\
\bottomrule
\end{tabular}
\end{table}

\begin{table}[h]
\addtolength{\tabcolsep}{-4pt}
\centering
\caption{Classification accuracy (mean $\pm$ std) for the Possum Age dataset (Part 2).}
\begin{tabular}{c|cccccccc}
 \toprule
\diagbox{BR}{Model} & RF[mss6] & RF[mss8] & RF[msl2] & RF[msl3] & RF[msl4] & RF[msl5] & RF[mfLog2] & RF[mfAll] \\
 \midrule
0.2 & 0.893 $\pm$ 0.107 & 0.902 $\pm$ 0.110 & 0.870 $\pm$ 0.109 & 0.867 $\pm$ 0.112 & 0.879 $\pm$ 0.118 & 0.969 $\pm$ 0.124 & 0.885 $\pm$ 0.102 & 0.889 $\pm$ 0.108 \\
0.4 & 0.855 $\pm$ 0.104 & 0.854 $\pm$ 0.105 & 0.848 $\pm$ 0.102 & 0.834 $\pm$ 0.104 & 0.828 $\pm$ 0.106 & 0.828 $\pm$ 0.109 & 0.856 $\pm$ 0.103 & 0.856 $\pm$ 0.102 \\
0.6 & 0.860 $\pm$ 0.108 & 0.854 $\pm$ 0.108 & 0.853 $\pm$ 0.108 & 0.843 $\pm$ 0.107 & 0.829 $\pm$ 0.107 & 0.818 $\pm$ 0.110 & 0.851 $\pm$ 0.102 & 0.864 $\pm$ 0.106 \\
0.8 & 0.870 $\pm$ 0.114 & 0.864 $\pm$ 0.113 & 0.865 $\pm$ 0.112 & 0.852 $\pm$ 0.110 & 0.839 $\pm$ 0.107 & 0.829 $\pm$ 0.109 & 0.849 $\pm$ 0.099 & 0.876 $\pm$ 0.113 \\
1 & 0.883 $\pm$ 0.116 & 0.875 $\pm$ 0.113 & 0.878 $\pm$ 0.114 & 0.866 $\pm$ 0.110 & 0.854 $\pm$ 0.109 & 0.840 $\pm$ 0.111 & 0.853 $\pm$ 0.107 & 0.890 $\pm$ 0.116 \\
1.2 & 0.905 $\pm$ 0.120 & 0.901 $\pm$ 0.119 & 0.896 $\pm$ 0.117 & 0.887 $\pm$ 0.115 & 0.876 $\pm$ 0.111 & 0.867 $\pm$ 0.113 & 0.862 $\pm$ 0.105 & 0.909 $\pm$ 0.120 \\
2 & 0.969 $\pm$ 0.138 & 0.965 $\pm$ 0.139 & 0.956 $\pm$ 0.135 & 0.945 $\pm$ 0.135 & 0.933 $\pm$ 0.128 & 0.922 $\pm$ 0.126 & 0.882 $\pm$ 0.114 & 0.970 $\pm$ 0.134 \\
3 & 1.015 $\pm$ 0.153 & 1.012 $\pm$ 0.152 & 1.007 $\pm$ 0.153 & 0.994 $\pm$ 0.150 & 0.982 $\pm$ 0.149 & 0.969 $\pm$ 0.143 & 0.897 $\pm$ 0.115 & 1.015 $\pm$ 0.153 \\
4 & 1.048 $\pm$ 0.163 & 1.047 $\pm$ 0.162 & 1.043 $\pm$ 0.161 & 1.030 $\pm$ 0.158 & 1.019 $\pm$ 0.159 & 1.012 $\pm$ 0.159 & 0.902 $\pm$ 0.113 & 1.049 $\pm$ 0.163 \\
5 & 1.068 $\pm$ 0.175 & 1.069 $\pm$ 0.176 & 1.066 $\pm$ 0.174 & 1.058 $\pm$ 0.171 & 1.044 $\pm$ 0.166 & 1.040 $\pm$ 0.168 & 0.907 $\pm$ 0.118 & 1.068 $\pm$ 0.174 \\
 \bottomrule
\end{tabular}
\end{table}

\begin{table}[h]
\addtolength{\tabcolsep}{-4pt}
\centering
\caption{Classification accuracy (mean $\pm$ std) for the Boston House Prices dataset (Part 1).}
\begin{tabular}{c|ccccccccc}
 \toprule
\diagbox{BR}{Model} & RF[100] & RF[200] & RF[500] & RF[md10] & RF[md15] & RF[md20] & RF[md25] & RF[mss3] & RF[mss4] \\
 \midrule
0.2 & 0.206 $\pm$ 0.040 & 0.204 $\pm$ 0.039 & 0.202 $\pm$ 0.038 & 0.206 $\pm$ 0.039 & 0.206 $\pm$ 0.040 & 0.206 $\pm$ 0.040 & 0.206 $\pm$ 0.040 & 0.206 $\pm$ 0.040 & 0.207 $\pm$ 0.040 \\
0.4 & 0.178 $\pm$ 0.038 & 0.176 $\pm$ 0.037 & 0.175 $\pm$ 0.036 & 0.178 $\pm$ 0.038 & 0.178 $\pm$ 0.038 & 0.178 $\pm$ 0.038 & 0.178 $\pm$ 0.038 & 0.178 $\pm$ 0.038 & 0.179 $\pm$ 0.038 \\
0.6 & 0.166 $\pm$ 0.037 & 0.165 $\pm$ 0.037 & 0.164 $\pm$ 0.037 & 0.166 $\pm$ 0.037 & 0.166 $\pm$ 0.037 & 0.166 $\pm$ 0.037 & 0.166 $\pm$ 0.037 & 0.166 $\pm$ 0.037 & 0.167 $\pm$ 0.037 \\
0.8 & 0.161 $\pm$ 0.037 & 0.160 $\pm$ 0.037 & 0.159 $\pm$ 0.037 & 0.161 $\pm$ 0.037 & 0.161 $\pm$ 0.037 & 0.161 $\pm$ 0.037 & 0.161 $\pm$ 0.037 & 0.161 $\pm$ 0.036 & 0.162 $\pm$ 0.036 \\
1 & 0.158 $\pm$ 0.037 & 0.158 $\pm$ 0.038 & 0.157 $\pm$ 0.038 & 0.158 $\pm$ 0.037 & 0.158 $\pm$ 0.037 & 0.158 $\pm$ 0.037 & 0.158 $\pm$ 0.037 & 0.158 $\pm$ 0.038 & 0.159 $\pm$ 0.037 \\
1.2 & 0.158 $\pm$ 0.039 & 0.157 $\pm$ 0.039 & 0.156 $\pm$ 0.039 & 0.158 $\pm$ 0.039 & 0.158 $\pm$ 0.039 & 0.158 $\pm$ 0.039 & 0.158 $\pm$ 0.039 & 0.158 $\pm$ 0.039 & 0.159 $\pm$ 0.039 \\
2 & 0.161 $\pm$ 0.041 & 0.160 $\pm$ 0.041 & 0.160 $\pm$ 0.041 & 0.162 $\pm$ 0.040 & 0.161 $\pm$ 0.040 & 0.161 $\pm$ 0.041 & 0.161 $\pm$ 0.041 & 0.161 $\pm$ 0.040 & 0.161 $\pm$ 0.040 \\
\bottomrule
\end{tabular}
\end{table}

\clearpage

\begin{table}[t]
\addtolength{\tabcolsep}{-4pt}
\centering
\caption{Classification accuracy (mean $\pm$ std) for the Boston House Prices dataset (Part 2).}
\begin{tabular}{c|cccccccc}
 \toprule
\diagbox{BR}{Model} & RF[mss6] & RF[mss8] & RF[msl2] & RF[msl3] & RF[msl4] & RF[msl5] & RF[mfLog2] & RF[mfAll] \\
 \midrule
0.2 & 0.211 $\pm$ 0.040 & 0.219 $\pm$ 0.040 & 0.212 $\pm$ 0.039 & 0.223 $\pm$ 0.041 & 0.239 $\pm$ 0.043 & 0.257 $\pm$ 0.045 & 0.250 $\pm$ 0.052 & 0.206 $\pm$ 0.040 \\
0.4 & 0.181 $\pm$ 0.038 & 0.184 $\pm$ 0.038 & 0.188 $\pm$ 0.038 & 0.195 $\pm$ 0.038 & 0.202 $\pm$ 0.038 & 0.212 $\pm$ 0.039 & 0.201 $\pm$ 0.044 & 0.178 $\pm$ 0.038 \\
0.6 & 0.168 $\pm$ 0.037 & 0.170 $\pm$ 0.037 & 0.179 $\pm$ 0.037 & 0.187 $\pm$ 0.037 & 0.193 $\pm$ 0.037 & 0.199 $\pm$ 0.037 & 0.177 $\pm$ 0.038 & 0.166 $\pm$ 0.037 \\
0.8 & 0.163 $\pm$ 0.036 & 0.164 $\pm$ 0.036 & 0.173 $\pm$ 0.037 & 0.183 $\pm$ 0.037 & 0.190 $\pm$ 0.037 & 0.195 $\pm$ 0.037 & 0.167 $\pm$ 0.039 & 0.161 $\pm$ 0.037 \\
1 & 0.160 $\pm$ 0.037 & 0.161 $\pm$ 0.038 & 0.170 $\pm$ 0.038 & 0.181 $\pm$ 0.038 & 0.188 $\pm$ 0.037 & 0.193 $\pm$ 0.037 & 0.160 $\pm$ 0.037 & 0.158 $\pm$ 0.037 \\
1.2 & 0.159 $\pm$ 0.039 & 0.160 $\pm$ 0.038 & 0.165 $\pm$ 0.039 & 0.174 $\pm$ 0.038 & 0.182 $\pm$ 0.039 & 0.187 $\pm$ 0.038 & 0.156 $\pm$ 0.037 & 0.158 $\pm$ 0.039 \\
2 & 0.162 $\pm$ 0.040 & 0.163 $\pm$ 0.040 & 0.165 $\pm$ 0.040 & 0.172 $\pm$ 0.040 & 0.179 $\pm$ 0.040 & 0.185 $\pm$ 0.040 & 0.147 $\pm$ 0.035 & 0.161 $\pm$ 0.041 \\
3 & 0.170 $\pm$ 0.044 & 0.170 $\pm$ 0.043 & 0.169 $\pm$ 0.043 & 0.172 $\pm$ 0.042 & 0.176 $\pm$ 0.042 & 0.180 $\pm$ 0.041 & 0.143 $\pm$ 0.033 & 0.169 $\pm$ 0.044 \\
4 & 0.175 $\pm$ 0.047 & 0.176 $\pm$ 0.047 & 0.175 $\pm$ 0.047 & 0.175 $\pm$ 0.045 & 0.177 $\pm$ 0.044 & 0.180 $\pm$ 0.043 & 0.142 $\pm$ 0.034 & 0.175 $\pm$ 0.047 \\
5 & 0.181 $\pm$ 0.049 & 0.180 $\pm$ 0.048 & 0.180 $\pm$ 0.048 & 0.179 $\pm$ 0.047 & 0.180 $\pm$ 0.046 & 0.182 $\pm$ 0.045 & 0.143 $\pm$ 0.035 & 0.180 $\pm$ 0.049 \\
 \bottomrule
\end{tabular}
\end{table}

\begin{table}[t]
\addtolength{\tabcolsep}{-4pt}
\centering
\caption{Classification accuracy (mean $\pm$ std) for the Energy Efficiency dataset (Part 1).}
\begin{tabular}{c|ccccccccc}
 \toprule
\diagbox{BR}{Model} & RF[100] & RF[200] & RF[500] & RF[md10] & RF[md15] & RF[md20] & RF[md25] & RF[mss3] & RF[mss4] \\
 \midrule
0.2 & 0.011 $\pm$ 0.003 & 0.011 $\pm$ 0.003 & 0.011 $\pm$ 0.003 & 0.011 $\pm$ 0.003 & 0.011 $\pm$ 0.003 & 0.011 $\pm$ 0.003 & 0.011 $\pm$ 0.003 & 0.011 $\pm$ 0.003 & 0.013 $\pm$ 0.003 \\
0.4 & 0.005 $\pm$ 0.001 & 0.005 $\pm$ 0.002 & 0.005 $\pm$ 0.002 & 0.005 $\pm$ 0.001 & 0.005 $\pm$ 0.001 & 0.005 $\pm$ 0.001 & 0.005 $\pm$ 0.001 & 0.005 $\pm$ 0.001 & 0.005 $\pm$ 0.001 \\
0.6 & 0.004 $\pm$ 0.001 & 0.004 $\pm$ 0.001 & 0.004 $\pm$ 0.001 & 0.004 $\pm$ 0.001 & 0.004 $\pm$ 0.001 & 0.004 $\pm$ 0.001 & 0.004 $\pm$ 0.001 & 0.004 $\pm$ 0.001 & 0.004 $\pm$ 0.001 \\
0.8 & 0.004 $\pm$ 0.001 & 0.004 $\pm$ 0.001 & 0.004 $\pm$ 0.001 & 0.004 $\pm$ 0.001 & 0.004 $\pm$ 0.001 & 0.004 $\pm$ 0.001 & 0.004 $\pm$ 0.001 & 0.004 $\pm$ 0.001 & 0.004 $\pm$ 0.001 \\
1 & 0.004 $\pm$ 0.001 & 0.004 $\pm$ 0.001 & 0.004 $\pm$ 0.001 & 0.004 $\pm$ 0.001 & 0.004 $\pm$ 0.001 & 0.004 $\pm$ 0.001 & 0.004 $\pm$ 0.001 & 0.004 $\pm$ 0.001 & 0.004 $\pm$ 0.001 \\
1.2 & 0.003 $\pm$ 0.001 & 0.003 $\pm$ 0.001 & 0.003 $\pm$ 0.001 & 0.003 $\pm$ 0.001 & 0.003 $\pm$ 0.001 & 0.003 $\pm$ 0.001 & 0.003 $\pm$ 0.001 & 0.004 $\pm$ 0.001 & 0.004 $\pm$ 0.001 \\
2 & 0.004 $\pm$ 0.001 & 0.004 $\pm$ 0.001 & 0.004 $\pm$ 0.001 & 0.004 $\pm$ 0.001 & 0.004 $\pm$ 0.001 & 0.004 $\pm$ 0.001 & 0.004 $\pm$ 0.001 & 0.004 $\pm$ 0.001 & 0.004 $\pm$ 0.001 \\
\bottomrule
\end{tabular}
\end{table}

\begin{table}[t]
\addtolength{\tabcolsep}{-4pt}
\centering
\caption{Classification accuracy (mean $\pm$ std) for the Energy Efficiency dataset (Part 2).}
\begin{tabular}{c|cccccccc}
 \toprule
\diagbox{BR}{Model} & RF[mss6] & RF[mss8] & RF[msl2] & RF[msl3] & RF[msl4] & RF[msl5] & RF[mfLog2] & RF[mfAll] \\
 \midrule
0.2 & 0.017 $\pm$ 0.003 & 0.021 $\pm$ 0.004 & 0.019 $\pm$ 0.005 & 0.028 $\pm$ 0.006 & 0.037 $\pm$ 0.006 & 0.045 $\pm$ 0.006 & 0.016 $\pm$ 0.004 & 0.011 $\pm$ 0.003 \\
0.4 & 0.006 $\pm$ 0.002 & 0.009 $\pm$ 0.002 & 0.008 $\pm$ 0.003 & 0.014 $\pm$ 0.004 & 0.021 $\pm$ 0.005 & 0.027 $\pm$ 0.005 & 0.008 $\pm$ 0.002 & 0.005 $\pm$ 0.001 \\
0.6 & 0.005 $\pm$ 0.001 & 0.005 $\pm$ 0.001 & 0.006 $\pm$ 0.002 & 0.009 $\pm$ 0.003 & 0.014 $\pm$ 0.004 & 0.020 $\pm$ 0.005 & 0.006 $\pm$ 0.002 & 0.004 $\pm$ 0.001 \\
0.8 & 0.004 $\pm$ 0.001 & 0.005 $\pm$ 0.001 & 0.005 $\pm$ 0.002 & 0.007 $\pm$ 0.003 & 0.011 $\pm$ 0.004 & 0.015 $\pm$ 0.005 & 0.005 $\pm$ 0.002 & 0.004 $\pm$ 0.001 \\
1 & 0.004 $\pm$ 0.001 & 0.004 $\pm$ 0.001 & 0.004 $\pm$ 0.001 & 0.006 $\pm$ 0.003 & 0.009 $\pm$ 0.004 & 0.013 $\pm$ 0.004 & 0.005 $\pm$ 0.002 & 0.004 $\pm$ 0.001 \\
1.2 & 0.004 $\pm$ 0.001 & 0.004 $\pm$ 0.001 & 0.004 $\pm$ 0.001 & 0.005 $\pm$ 0.002 & 0.006 $\pm$ 0.003 & 0.008 $\pm$ 0.003 & 0.005 $\pm$ 0.002 & 0.003 $\pm$ 0.001 \\
2 & 0.004 $\pm$ 0.001 & 0.004 $\pm$ 0.001 & 0.004 $\pm$ 0.001 & 0.004 $\pm$ 0.001 & 0.005 $\pm$ 0.002 & 0.006 $\pm$ 0.003 & 0.004 $\pm$ 0.001 & 0.004 $\pm$ 0.001 \\
3 & 0.004 $\pm$ 0.001 & 0.004 $\pm$ 0.001 & 0.004 $\pm$ 0.001 & 0.004 $\pm$ 0.001 & 0.004 $\pm$ 0.001 & 0.005 $\pm$ 0.002 & 0.004 $\pm$ 0.002 & 0.004 $\pm$ 0.001 \\
4 & 0.004 $\pm$ 0.001 & 0.004 $\pm$ 0.001 & 0.004 $\pm$ 0.001 & 0.004 $\pm$ 0.001 & 0.004 $\pm$ 0.001 & 0.004 $\pm$ 0.001 & 0.004 $\pm$ 0.001 & 0.004 $\pm$ 0.001 \\
5 & 0.004 $\pm$ 0.001 & 0.004 $\pm$ 0.001 & 0.004 $\pm$ 0.001 & 0.004 $\pm$ 0.001 & 0.004 $\pm$ 0.001 & 0.004 $\pm$ 0.001 & 0.004 $\pm$ 0.002 & 0.004 $\pm$ 0.001 \\
 \bottomrule
\end{tabular}
\end{table}

\begin{table}[t]
\addtolength{\tabcolsep}{-4pt}
\centering
\caption{Classification accuracy (mean $\pm$ std) for the Abalone dataset (Part 1).}
\begin{tabular}{c|ccccccccc}
 \toprule
\diagbox{BR}{Model} & RF[100] & RF[200] & RF[500] & RF[md10] & RF[md15] & RF[md20] & RF[md25] & RF[mss3] & RF[mss4] \\
 \midrule
0.2 & 0.446 $\pm$ 0.017 & 0.443 $\pm$ 0.018 & 0.442 $\pm$ 0.018 & 0.446 $\pm$ 0.018 & 0.446 $\pm$ 0.017 & 0.446 $\pm$ 0.017 & 0.446 $\pm$ 0.017 & 0.445 $\pm$ 0.018 & 0.445 $\pm$ 0.018 \\
0.4 & 0.448 $\pm$ 0.017 & 0.446 $\pm$ 0.017 & 0.445 $\pm$ 0.017 & 0.447 $\pm$ 0.017 & 0.448 $\pm$ 0.017 & 0.448 $\pm$ 0.017 & 0.448 $\pm$ 0.017 & 0.448 $\pm$ 0.017 & 0.447 $\pm$ 0.017 \\
0.6 & 0.454 $\pm$ 0.016 & 0.452 $\pm$ 0.016 & 0.451 $\pm$ 0.016 & 0.451 $\pm$ 0.017 & 0.454 $\pm$ 0.017 & 0.454 $\pm$ 0.016 & 0.454 $\pm$ 0.016 & 0.454 $\pm$ 0.016 & 0.453 $\pm$ 0.016 \\
0.8 & 0.460 $\pm$ 0.017 & 0.458 $\pm$ 0.017 & 0.456 $\pm$ 0.016 & 0.455 $\pm$ 0.017 & 0.459 $\pm$ 0.016 & 0.460 $\pm$ 0.017 & 0.460 $\pm$ 0.017 & 0.459 $\pm$ 0.016 & 0.458 $\pm$ 0.016 \\
1 & 0.465 $\pm$ 0.016 & 0.463 $\pm$ 0.016 & 0.462 $\pm$ 0.016 & 0.459 $\pm$ 0.017 & 0.464 $\pm$ 0.016 & 0.465 $\pm$ 0.016 & 0.465 $\pm$ 0.016 & 0.465 $\pm$ 0.016 & 0.464 $\pm$ 0.016 \\
1.2 & 0.470 $\pm$ 0.016 & 0.469 $\pm$ 0.016 & 0.467 $\pm$ 0.016 & 0.462 $\pm$ 0.017 & 0.469 $\pm$ 0.016 & 0.470 $\pm$ 0.016 & 0.470 $\pm$ 0.016 & 0.470 $\pm$ 0.016 & 0.469 $\pm$ 0.016 \\
2 & 0.486 $\pm$ 0.017 & 0.484 $\pm$ 0.017 & 0.483 $\pm$ 0.017 & 0.473 $\pm$ 0.017 & 0.484 $\pm$ 0.017 & 0.486 $\pm$ 0.017 & 0.486 $\pm$ 0.017 & 0.486 $\pm$ 0.017 & 0.485 $\pm$ 0.017 \\
\bottomrule
\end{tabular}
\end{table}

\begin{table}[t]
\addtolength{\tabcolsep}{-4pt}
\centering
\caption{Classification accuracy (mean $\pm$ std) for the Abalone dataset (Part 2).}
\begin{tabular}{c|cccccccc}
 \toprule
\diagbox{BR}{Model} & RF[mss6] & RF[mss8] & RF[msl2] & RF[msl3] & RF[msl4] & RF[msl5] & RF[mfLog2] & RF[mfAll] \\
 \midrule
0.2 & 0.445 $\pm$ 0.018 & 0.445 $\pm$ 0.018 & 0.445 $\pm$ 0.018 & 0.445 $\pm$ 0.018 & 0.445 $\pm$ 0.019 & 0.447 $\pm$ 0.019 & 0.458 $\pm$ 0.020 & 0.446 $\pm$ 0.017 \\
0.4 & 0.447 $\pm$ 0.017 & 0.446 $\pm$ 0.017 & 0.446 $\pm$ 0.017 & 0.445 $\pm$ 0.017 & 0.444 $\pm$ 0.018 & 0.444 $\pm$ 0.018 & 0.452 $\pm$ 0.018 & 0.448 $\pm$ 0.017 \\
0.6 & 0.452 $\pm$ 0.016 & 0.450 $\pm$ 0.017 & 0.451 $\pm$ 0.017 & 0.448 $\pm$ 0.017 & 0.447 $\pm$ 0.017 & 0.446 $\pm$ 0.017 & 0.452 $\pm$ 0.018 & 0.454 $\pm$ 0.016 \\
0.8 & 0.456 $\pm$ 0.017 & 0.455 $\pm$ 0.017 & 0.455 $\pm$ 0.017 & 0.453 $\pm$ 0.017 & 0.451 $\pm$ 0.017 & 0.449 $\pm$ 0.017 & 0.454 $\pm$ 0.018 & 0.460 $\pm$ 0.017 \\
1 & 0.461 $\pm$ 0.016 & 0.459 $\pm$ 0.016 & 0.460 $\pm$ 0.017 & 0.456 $\pm$ 0.017 & 0.454 $\pm$ 0.017 & 0.452 $\pm$ 0.017 & 0.456 $\pm$ 0.018 & 0.465 $\pm$ 0.016 \\
1.2 & 0.467 $\pm$ 0.016 & 0.465 $\pm$ 0.016 & 0.465 $\pm$ 0.017 & 0.462 $\pm$ 0.017 & 0.459 $\pm$ 0.017 & 0.456 $\pm$ 0.017 & 0.458 $\pm$ 0.017 & 0.470 $\pm$ 0.016 \\
2 & 0.483 $\pm$ 0.017 & 0.481 $\pm$ 0.017 & 0.480 $\pm$ 0.017 & 0.475 $\pm$ 0.017 & 0.471 $\pm$ 0.017 & 0.468 $\pm$ 0.017 & 0.462 $\pm$ 0.017 & 0.486 $\pm$ 0.017 \\
3 & 0.497 $\pm$ 0.017 & 0.495 $\pm$ 0.017 & 0.494 $\pm$ 0.017 & 0.489 $\pm$ 0.017 & 0.485 $\pm$ 0.017 & 0.481 $\pm$ 0.017 & 0.465 $\pm$ 0.016 & 0.499 $\pm$ 0.017 \\
4 & 0.507 $\pm$ 0.018 & 0.506 $\pm$ 0.018 & 0.505 $\pm$ 0.018 & 0.499 $\pm$ 0.018 & 0.495 $\pm$ 0.018 & 0.492 $\pm$ 0.018 & 0.466 $\pm$ 0.017 & 0.508 $\pm$ 0.018 \\
5 & 0.514 $\pm$ 0.018 & 0.513 $\pm$ 0.018 & 0.513 $\pm$ 0.018 & 0.509 $\pm$ 0.018 & 0.504 $\pm$ 0.018 & 0.501 $\pm$ 0.018 & 0.468 $\pm$ 0.017 & 0.515 $\pm$ 0.018 \\
 \bottomrule
\end{tabular}
\end{table}

\begin{table}[t]
\addtolength{\tabcolsep}{-4pt}
\centering
\caption{Classification accuracy (mean $\pm$ std) for the Taxi price dataset (Part 1).}
\begin{tabular}{c|ccccccccc}
 \toprule
\diagbox{BR}{Model} & RF[100] & RF[200] & RF[500] & RF[md10] & RF[md15] & RF[md20] & RF[md25] & RF[mss3] & RF[mss4] \\
 \midrule
0.2 & 0.098 $\pm$ 0.026 & 0.098 $\pm$ 0.027 & 0.097 $\pm$ 0.025 & 0.097 $\pm$ 0.026 & 0.098 $\pm$ 0.026 & 0.098 $\pm$ 0.026 & 0.098 $\pm$ 0.026 & 0.098 $\pm$ 0.026 & 0.099 $\pm$ 0.026 \\
0.4 & 0.075 $\pm$ 0.011 & 0.074 $\pm$ 0.011 & 0.074 $\pm$ 0.011 & 0.075 $\pm$ 0.011 & 0.075 $\pm$ 0.011 & 0.075 $\pm$ 0.011 & 0.075 $\pm$ 0.011 & 0.075 $\pm$ 0.011 & 0.076 $\pm$ 0.011 \\
0.6 & 0.071 $\pm$ 0.011 & 0.071 $\pm$ 0.011 & 0.071 $\pm$ 0.011 & 0.071 $\pm$ 0.011 & 0.072 $\pm$ 0.011 & 0.071 $\pm$ 0.011 & 0.071 $\pm$ 0.011 & 0.072 $\pm$ 0.012 & 0.072 $\pm$ 0.011 \\
0.8 & 0.071 $\pm$ 0.012 & 0.070 $\pm$ 0.012 & 0.070 $\pm$ 0.012 & 0.071 $\pm$ 0.012 & 0.071 $\pm$ 0.012 & 0.071 $\pm$ 0.012 & 0.071 $\pm$ 0.012 & 0.071 $\pm$ 0.012 & 0.071 $\pm$ 0.012 \\
1 & 0.071 $\pm$ 0.013 & 0.070 $\pm$ 0.012 & 0.070 $\pm$ 0.012 & 0.071 $\pm$ 0.012 & 0.071 $\pm$ 0.012 & 0.071 $\pm$ 0.013 & 0.071 $\pm$ 0.013 & 0.071 $\pm$ 0.013 & 0.071 $\pm$ 0.013 \\
1.2 & 0.071 $\pm$ 0.013 & 0.071 $\pm$ 0.013 & 0.071 $\pm$ 0.013 & 0.072 $\pm$ 0.013 & 0.072 $\pm$ 0.014 & 0.071 $\pm$ 0.013 & 0.071 $\pm$ 0.013 & 0.071 $\pm$ 0.014 & 0.071 $\pm$ 0.013 \\
2 & 0.075 $\pm$ 0.016 & 0.074 $\pm$ 0.016 & 0.074 $\pm$ 0.016 & 0.075 $\pm$ 0.016 & 0.075 $\pm$ 0.016 & 0.075 $\pm$ 0.016 & 0.075 $\pm$ 0.016 & 0.075 $\pm$ 0.016 & 0.075 $\pm$ 0.016 \\
\bottomrule
\end{tabular}
\end{table}

\begin{table}[t]
\addtolength{\tabcolsep}{-4pt}
\centering
\caption{Classification accuracy (mean $\pm$ std) for the Taxi price dataset (Part 2).}
\begin{tabular}{c|cccccccc}
 \toprule
\diagbox{BR}{Model} & RF[mss6] & RF[mss8] & RF[msl2] & RF[msl3] & RF[msl4] & RF[msl5] & RF[mfLog2] & RF[mfAll] \\
 \midrule
0.2 & 0.101 $\pm$ 0.026 & 0.103 $\pm$ 0.026 & 0.146 $\pm$ 0.057 & 0.228 $\pm$ 0.085 & 0.314 $\pm$ 0.095 & 0.380 $\pm$ 0.095 & 0.429 $\pm$ 0.092 & 0.098 $\pm$ 0.026 \\
0.4 & 0.077 $\pm$ 0.011 & 0.079 $\pm$ 0.011 & 0.080 $\pm$ 0.019 & 0.099 $\pm$ 0.036 & 0.142 $\pm$ 0.062 & 0.205 $\pm$ 0.083 & 0.330 $\pm$ 0.074 & 0.075 $\pm$ 0.011 \\
0.6 & 0.073 $\pm$ 0.011 & 0.075 $\pm$ 0.011 & 0.072 $\pm$ 0.012 & 0.076 $\pm$ 0.016 & 0.091 $\pm$ 0.033 & 0.123 $\pm$ 0.057 & 0.287 $\pm$ 0.064 & 0.071 $\pm$ 0.011 \\
0.8 & 0.072 $\pm$ 0.011 & 0.074 $\pm$ 0.011 & 0.071 $\pm$ 0.012 & 0.072 $\pm$ 0.012 & 0.078 $\pm$ 0.018 & 0.092 $\pm$ 0.035 & 0.264 $\pm$ 0.058 & 0.071 $\pm$ 0.012 \\
1 & 0.072 $\pm$ 0.012 & 0.073 $\pm$ 0.011 & 0.071 $\pm$ 0.013 & 0.071 $\pm$ 0.011 & 0.074 $\pm$ 0.012 & 0.080 $\pm$ 0.021 & 0.250 $\pm$ 0.058 & 0.071 $\pm$ 0.013 \\
1.2 & 0.072 $\pm$ 0.013 & 0.072 $\pm$ 0.012 & 0.071 $\pm$ 0.013 & 0.072 $\pm$ 0.012 & 0.072 $\pm$ 0.011 & 0.073 $\pm$ 0.011 & 0.242 $\pm$ 0.052 & 0.071 $\pm$ 0.013 \\
2 & 0.075 $\pm$ 0.016 & 0.075 $\pm$ 0.016 & 0.075 $\pm$ 0.016 & 0.075 $\pm$ 0.016 & 0.074 $\pm$ 0.014 & 0.074 $\pm$ 0.012 & 0.225 $\pm$ 0.052 & 0.075 $\pm$ 0.016 \\
3 & 0.078 $\pm$ 0.017 & 0.079 $\pm$ 0.018 & 0.078 $\pm$ 0.017 & 0.078 $\pm$ 0.017 & 0.078 $\pm$ 0.017 & 0.077 $\pm$ 0.016 & 0.220 $\pm$ 0.050 & 0.078 $\pm$ 0.017 \\
4 & 0.081 $\pm$ 0.018 & 0.081 $\pm$ 0.018 & 0.081 $\pm$ 0.018 & 0.081 $\pm$ 0.018 & 0.081 $\pm$ 0.018 & 0.080 $\pm$ 0.018 & 0.215 $\pm$ 0.051 & 0.081 $\pm$ 0.018 \\
5 & 0.083 $\pm$ 0.019 & 0.083 $\pm$ 0.019 & 0.083 $\pm$ 0.019 & 0.082 $\pm$ 0.019 & 0.082 $\pm$ 0.019 & 0.083 $\pm$ 0.019 & 0.214 $\pm$ 0.051 & 0.083 $\pm$ 0.018 \\
 \bottomrule
\end{tabular}
\end{table}

\begin{table}[t]
\addtolength{\tabcolsep}{-4pt}
\centering
\caption{Classification accuracy (mean $\pm$ std) for the Food Delivery Time dataset (Part 1).}
\begin{tabular}{c|ccccccccc}
 \toprule
\diagbox{BR}{Model} & RF[100] & RF[200] & RF[500] & RF[md10] & RF[md15] & RF[md20] & RF[md25] & RF[mss3] & RF[mss4] \\
 \midrule
0.2 & 0.278 $\pm$ 0.027 & 0.276 $\pm$ 0.027 & 0.275 $\pm$ 0.027 & 0.278 $\pm$ 0.027 & 0.278 $\pm$ 0.027 & 0.278 $\pm$ 0.027 & 0.278 $\pm$ 0.027 & 0.278 $\pm$ 0.027 & 0.278 $\pm$ 0.027 \\
0.4 & 0.275 $\pm$ 0.027 & 0.274 $\pm$ 0.027 & 0.273 $\pm$ 0.027 & 0.275 $\pm$ 0.027 & 0.275 $\pm$ 0.027 & 0.275 $\pm$ 0.027 & 0.275 $\pm$ 0.027 & 0.275 $\pm$ 0.027 & 0.275 $\pm$ 0.027 \\
0.6 & 0.278 $\pm$ 0.026 & 0.277 $\pm$ 0.026 & 0.276 $\pm$ 0.026 & 0.278 $\pm$ 0.026 & 0.278 $\pm$ 0.026 & 0.278 $\pm$ 0.026 & 0.278 $\pm$ 0.026 & 0.277 $\pm$ 0.026 & 0.277 $\pm$ 0.026 \\
0.8 & 0.284 $\pm$ 0.026 & 0.282 $\pm$ 0.027 & 0.281 $\pm$ 0.027 & 0.284 $\pm$ 0.026 & 0.284 $\pm$ 0.026 & 0.284 $\pm$ 0.026 & 0.284 $\pm$ 0.026 & 0.283 $\pm$ 0.026 & 0.282 $\pm$ 0.027 \\
1 & 0.288 $\pm$ 0.027 & 0.287 $\pm$ 0.027 & 0.286 $\pm$ 0.027 & 0.288 $\pm$ 0.027 & 0.288 $\pm$ 0.027 & 0.289 $\pm$ 0.027 & 0.288 $\pm$ 0.027 & 0.287 $\pm$ 0.027 & 0.286 $\pm$ 0.027 \\
1.2 & 0.293 $\pm$ 0.027 & 0.291 $\pm$ 0.027 & 0.291 $\pm$ 0.027 & 0.293 $\pm$ 0.027 & 0.293 $\pm$ 0.027 & 0.293 $\pm$ 0.027 & 0.293 $\pm$ 0.027 & 0.293 $\pm$ 0.028 & 0.292 $\pm$ 0.028 \\
2 & 0.309 $\pm$ 0.028 & 0.307 $\pm$ 0.028 & 0.307 $\pm$ 0.028 & 0.308 $\pm$ 0.028 & 0.308 $\pm$ 0.028 & 0.309 $\pm$ 0.028 & 0.309 $\pm$ 0.028 & 0.308 $\pm$ 0.028 & 0.307 $\pm$ 0.028 \\
\bottomrule
\end{tabular}
\end{table}

\begin{table}[t]
\addtolength{\tabcolsep}{-4pt}
\centering
\caption{Classification accuracy (mean $\pm$ std) for the Food Delivery Time dataset (Part 2).}
\begin{tabular}{c|cccccccc}
 \toprule
\diagbox{BR}{Model} & RF[mss6] & RF[mss8] & RF[msl2] & RF[msl3] & RF[msl4] & RF[msl5] & RF[mfLog2] & RF[mfAll] \\
 \midrule
0.2 & 0.280 $\pm$ 0.027 & 0.283 $\pm$ 0.027 & 0.278 $\pm$ 0.027 & 0.281 $\pm$ 0.028 & 0.286 $\pm$ 0.028 & 0.291 $\pm$ 0.028 & 0.369 $\pm$ 0.037 & 0.278 $\pm$ 0.027 \\
0.4 & 0.275 $\pm$ 0.027 & 0.277 $\pm$ 0.027 & 0.273 $\pm$ 0.027 & 0.275 $\pm$ 0.027 & 0.278 $\pm$ 0.027 & 0.281 $\pm$ 0.027 & 0.337 $\pm$ 0.033 & 0.275 $\pm$ 0.027 \\
0.6 & 0.277 $\pm$ 0.026 & 0.277 $\pm$ 0.027 & 0.274 $\pm$ 0.026 & 0.275 $\pm$ 0.026 & 0.277 $\pm$ 0.027 & 0.279 $\pm$ 0.027 & 0.329 $\pm$ 0.034 & 0.278 $\pm$ 0.026 \\
0.8 & 0.281 $\pm$ 0.027 & 0.280 $\pm$ 0.027 & 0.278 $\pm$ 0.026 & 0.277 $\pm$ 0.026 & 0.278 $\pm$ 0.026 & 0.280 $\pm$ 0.027 & 0.324 $\pm$ 0.033 & 0.284 $\pm$ 0.026 \\
1 & 0.284 $\pm$ 0.027 & 0.284 $\pm$ 0.027 & 0.281 $\pm$ 0.027 & 0.279 $\pm$ 0.027 & 0.279 $\pm$ 0.027 & 0.280 $\pm$ 0.027 & 0.321 $\pm$ 0.033 & 0.288 $\pm$ 0.027 \\
1.2 & 0.289 $\pm$ 0.028 & 0.288 $\pm$ 0.028 & 0.286 $\pm$ 0.028 & 0.282 $\pm$ 0.027 & 0.281 $\pm$ 0.027 & 0.280 $\pm$ 0.027 & 0.320 $\pm$ 0.032 & 0.293 $\pm$ 0.027 \\
2 & 0.304 $\pm$ 0.028 & 0.301 $\pm$ 0.028 & 0.298 $\pm$ 0.028 & 0.293 $\pm$ 0.027 & 0.290 $\pm$ 0.027 & 0.287 $\pm$ 0.027 & 0.322 $\pm$ 0.033 & 0.309 $\pm$ 0.028 \\
3 & 0.319 $\pm$ 0.029 & 0.316 $\pm$ 0.029 & 0.314 $\pm$ 0.028 & 0.305 $\pm$ 0.028 & 0.301 $\pm$ 0.028 & 0.298 $\pm$ 0.027 & 0.323 $\pm$ 0.032 & 0.322 $\pm$ 0.030 \\
4 & 0.329 $\pm$ 0.030 & 0.327 $\pm$ 0.030 & 0.326 $\pm$ 0.029 & 0.317 $\pm$ 0.028 & 0.311 $\pm$ 0.028 & 0.307 $\pm$ 0.028 & 0.324 $\pm$ 0.031 & 0.331 $\pm$ 0.030 \\
5 & 0.337 $\pm$ 0.031 & 0.336 $\pm$ 0.032 & 0.336 $\pm$ 0.031 & 0.328 $\pm$ 0.029 & 0.320 $\pm$ 0.028 & 0.316 $\pm$ 0.028 & 0.323 $\pm$ 0.032 & 0.339 $\pm$ 0.032 \\
 \bottomrule
\end{tabular}
\end{table}

\begin{table}[t]
\addtolength{\tabcolsep}{-4pt}
\centering
\caption{Classification accuracy (mean $\pm$ std) for the Airfoil Self-Noise dataset (Part 1).}
\begin{tabular}{c|ccccccccc}
 \toprule
\diagbox{BR}{Model} & RF[100] & RF[200] & RF[500] & RF[md10] & RF[md15] & RF[md20] & RF[md25] & RF[mss3] & RF[mss4] \\
 \midrule
0.2 & 0.188 $\pm$ 0.013 & 0.187 $\pm$ 0.012 & 0.185 $\pm$ 0.012 & 0.193 $\pm$ 0.013 & 0.188 $\pm$ 0.013 & 0.188 $\pm$ 0.013 & 0.188 $\pm$ 0.013 & 0.193 $\pm$ 0.013 & 0.202 $\pm$ 0.013 \\
0.4 & 0.134 $\pm$ 0.011 & 0.133 $\pm$ 0.011 & 0.131 $\pm$ 0.011 & 0.142 $\pm$ 0.011 & 0.134 $\pm$ 0.011 & 0.134 $\pm$ 0.011 & 0.134 $\pm$ 0.011 & 0.139 $\pm$ 0.011 & 0.146 $\pm$ 0.011 \\
0.6 & 0.113 $\pm$ 0.010 & 0.112 $\pm$ 0.010 & 0.111 $\pm$ 0.010 & 0.124 $\pm$ 0.010 & 0.113 $\pm$ 0.010 & 0.113 $\pm$ 0.010 & 0.113 $\pm$ 0.010 & 0.117 $\pm$ 0.010 & 0.124 $\pm$ 0.011 \\
0.8 & 0.103 $\pm$ 0.010 & 0.102 $\pm$ 0.010 & 0.101 $\pm$ 0.010 & 0.115 $\pm$ 0.010 & 0.103 $\pm$ 0.010 & 0.103 $\pm$ 0.010 & 0.103 $\pm$ 0.010 & 0.107 $\pm$ 0.010 & 0.114 $\pm$ 0.010 \\
1 & 0.098 $\pm$ 0.010 & 0.097 $\pm$ 0.010 & 0.096 $\pm$ 0.010 & 0.111 $\pm$ 0.010 & 0.098 $\pm$ 0.010 & 0.098 $\pm$ 0.010 & 0.098 $\pm$ 0.010 & 0.102 $\pm$ 0.010 & 0.108 $\pm$ 0.010 \\
1.2 & 0.094 $\pm$ 0.009 & 0.094 $\pm$ 0.010 & 0.093 $\pm$ 0.010 & 0.108 $\pm$ 0.010 & 0.095 $\pm$ 0.009 & 0.094 $\pm$ 0.009 & 0.094 $\pm$ 0.009 & 0.096 $\pm$ 0.009 & 0.100 $\pm$ 0.010 \\
2 & 0.092 $\pm$ 0.010 & 0.092 $\pm$ 0.010 & 0.091 $\pm$ 0.010 & 0.108 $\pm$ 0.011 & 0.093 $\pm$ 0.010 & 0.092 $\pm$ 0.010 & 0.092 $\pm$ 0.010 & 0.093 $\pm$ 0.010 & 0.096 $\pm$ 0.010 \\
\bottomrule
\end{tabular}
\end{table}

\begin{table}[t]
\addtolength{\tabcolsep}{-4pt}
\centering
\caption{Classification accuracy (mean $\pm$ std) for the Airfoil Self-Noise dataset (Part 2).}
\begin{tabular}{c|cccccccc}
 \toprule
\diagbox{BR}{Model} & RF[mss6] & RF[mss8] & RF[msl2] & RF[msl3] & RF[msl4] & RF[msl5] & RF[mfLog2] & RF[mfAll] \\
 \midrule
0.2 & 0.224 $\pm$ 0.014 & 0.247 $\pm$ 0.015 & 0.214 $\pm$ 0.014 & 0.245 $\pm$ 0.015 & 0.277 $\pm$ 0.016 & 0.305 $\pm$ 0.017 & 0.222 $\pm$ 0.012 & 0.188 $\pm$ 0.013 \\
0.4 & 0.164 $\pm$ 0.012 & 0.183 $\pm$ 0.013 & 0.156 $\pm$ 0.012 & 0.179 $\pm$ 0.013 & 0.203 $\pm$ 0.014 & 0.226 $\pm$ 0.015 & 0.160 $\pm$ 0.012 & 0.134 $\pm$ 0.011 \\
0.6 & 0.140 $\pm$ 0.012 & 0.157 $\pm$ 0.013 & 0.133 $\pm$ 0.011 & 0.153 $\pm$ 0.012 & 0.173 $\pm$ 0.013 & 0.193 $\pm$ 0.014 & 0.133 $\pm$ 0.011 & 0.113 $\pm$ 0.010 \\
0.8 & 0.128 $\pm$ 0.011 & 0.144 $\pm$ 0.012 & 0.122 $\pm$ 0.011 & 0.140 $\pm$ 0.012 & 0.159 $\pm$ 0.013 & 0.176 $\pm$ 0.014 & 0.120 $\pm$ 0.011 & 0.103 $\pm$ 0.010 \\
1 & 0.122 $\pm$ 0.011 & 0.137 $\pm$ 0.012 & 0.116 $\pm$ 0.010 & 0.133 $\pm$ 0.011 & 0.150 $\pm$ 0.012 & 0.166 $\pm$ 0.013 & 0.111 $\pm$ 0.011 & 0.098 $\pm$ 0.010 \\
1.2 & 0.110 $\pm$ 0.010 & 0.120 $\pm$ 0.011 & 0.107 $\pm$ 0.010 & 0.120 $\pm$ 0.010 & 0.133 $\pm$ 0.011 & 0.145 $\pm$ 0.012 & 0.107 $\pm$ 0.010 & 0.094 $\pm$ 0.009 \\
2 & 0.103 $\pm$ 0.010 & 0.111 $\pm$ 0.011 & 0.101 $\pm$ 0.010 & 0.112 $\pm$ 0.010 & 0.122 $\pm$ 0.011 & 0.132 $\pm$ 0.011 & 0.099 $\pm$ 0.012 & 0.092 $\pm$ 0.010 \\
3 & 0.101 $\pm$ 0.011 & 0.106 $\pm$ 0.011 & 0.099 $\pm$ 0.010 & 0.106 $\pm$ 0.011 & 0.114 $\pm$ 0.011 & 0.121 $\pm$ 0.011 & 0.098 $\pm$ 0.012 & 0.096 $\pm$ 0.011 \\
4 & 0.101 $\pm$ 0.011 & 0.104 $\pm$ 0.011 & 0.100 $\pm$ 0.011 & 0.104 $\pm$ 0.011 & 0.111 $\pm$ 0.011 & 0.116 $\pm$ 0.011 & 0.098 $\pm$ 0.012 & 0.099 $\pm$ 0.011 \\
5 & 0.103 $\pm$ 0.011 & 0.105 $\pm$ 0.011 & 0.103 $\pm$ 0.011 & 0.104 $\pm$ 0.011 & 0.109 $\pm$ 0.011 & 0.115 $\pm$ 0.012 & 0.099 $\pm$ 0.012 & 0.102 $\pm$ 0.011 \\
 \bottomrule
\end{tabular}
\end{table}

\begin{table}[t]
\addtolength{\tabcolsep}{-4pt}
\centering
\caption{Classification accuracy (mean $\pm$ std) for the Coffee Shop Revenue dataset (Part 1).}
\begin{tabular}{c|ccccccccc}
 \toprule
\diagbox{BR}{Model} & RF[100] & RF[200] & RF[500] & RF[md10] & RF[md15] & RF[md20] & RF[md25] & RF[mss3] & RF[mss4] \\
 \midrule
0.2 & 0.063 $\pm$ 0.003 & 0.062 $\pm$ 0.003 & 0.062 $\pm$ 0.003 & 0.063 $\pm$ 0.003 & 0.063 $\pm$ 0.003 & 0.063 $\pm$ 0.003 & 0.063 $\pm$ 0.003 & 0.063 $\pm$ 0.003 & 0.063 $\pm$ 0.003 \\
0.4 & 0.058 $\pm$ 0.003 & 0.057 $\pm$ 0.002 & 0.057 $\pm$ 0.002 & 0.058 $\pm$ 0.003 & 0.058 $\pm$ 0.003 & 0.058 $\pm$ 0.003 & 0.058 $\pm$ 0.003 & 0.058 $\pm$ 0.003 & 0.058 $\pm$ 0.003 \\
0.6 & 0.057 $\pm$ 0.002 & 0.056 $\pm$ 0.002 & 0.056 $\pm$ 0.002 & 0.057 $\pm$ 0.002 & 0.057 $\pm$ 0.002 & 0.057 $\pm$ 0.002 & 0.057 $\pm$ 0.002 & 0.057 $\pm$ 0.002 & 0.057 $\pm$ 0.002 \\
0.8 & 0.057 $\pm$ 0.002 & 0.057 $\pm$ 0.002 & 0.057 $\pm$ 0.002 & 0.057 $\pm$ 0.002 & 0.057 $\pm$ 0.002 & 0.057 $\pm$ 0.002 & 0.057 $\pm$ 0.002 & 0.057 $\pm$ 0.002 & 0.057 $\pm$ 0.002 \\
1 & 0.058 $\pm$ 0.002 & 0.057 $\pm$ 0.002 & 0.057 $\pm$ 0.002 & 0.058 $\pm$ 0.002 & 0.058 $\pm$ 0.002 & 0.058 $\pm$ 0.002 & 0.058 $\pm$ 0.002 & 0.058 $\pm$ 0.002 & 0.058 $\pm$ 0.002 \\
1.2 & 0.059 $\pm$ 0.002 & 0.058 $\pm$ 0.002 & 0.058 $\pm$ 0.002 & 0.058 $\pm$ 0.002 & 0.059 $\pm$ 0.002 & 0.059 $\pm$ 0.002 & 0.059 $\pm$ 0.002 & 0.058 $\pm$ 0.002 & 0.058 $\pm$ 0.002 \\
2 & 0.061 $\pm$ 0.002 & 0.061 $\pm$ 0.002 & 0.061 $\pm$ 0.002 & 0.061 $\pm$ 0.002 & 0.061 $\pm$ 0.002 & 0.061 $\pm$ 0.002 & 0.061 $\pm$ 0.002 & 0.061 $\pm$ 0.002 & 0.061 $\pm$ 0.002 \\
\bottomrule
\end{tabular}
\end{table}

\begin{table}[t]
\addtolength{\tabcolsep}{-4pt}
\centering
\caption{Classification accuracy (mean $\pm$ std) for the Coffee Shop Revenue dataset (Part 2).}
\begin{tabular}{c|cccccccc}
 \toprule
\diagbox{BR}{Model} & RF[mss6] & RF[mss8] & RF[msl2] & RF[msl3] & RF[msl4] & RF[msl5] & RF[mfLog2] & RF[mfAll] \\
 \midrule
0.2 & 0.064 $\pm$ 0.003 & 0.066 $\pm$ 0.003 & 0.063 $\pm$ 0.003 & 0.065 $\pm$ 0.003 & 0.068 $\pm$ 0.003 & 0.072 $\pm$ 0.004 & 0.108 $\pm$ 0.008 & 0.063 $\pm$ 0.003 \\
0.4 & 0.058 $\pm$ 0.003 & 0.059 $\pm$ 0.003 & 0.058 $\pm$ 0.003 & 0.058 $\pm$ 0.003 & 0.060 $\pm$ 0.003 & 0.062 $\pm$ 0.003 & 0.088 $\pm$ 0.005 & 0.058 $\pm$ 0.003 \\
0.6 & 0.057 $\pm$ 0.002 & 0.057 $\pm$ 0.002 & 0.057 $\pm$ 0.002 & 0.057 $\pm$ 0.002 & 0.058 $\pm$ 0.002 & 0.059 $\pm$ 0.003 & 0.080 $\pm$ 0.005 & 0.057 $\pm$ 0.002 \\
0.8 & 0.057 $\pm$ 0.002 & 0.057 $\pm$ 0.002 & 0.057 $\pm$ 0.002 & 0.057 $\pm$ 0.002 & 0.057 $\pm$ 0.002 & 0.058 $\pm$ 0.002 & 0.077 $\pm$ 0.005 & 0.057 $\pm$ 0.002 \\
1 & 0.058 $\pm$ 0.002 & 0.058 $\pm$ 0.002 & 0.057 $\pm$ 0.002 & 0.057 $\pm$ 0.002 & 0.057 $\pm$ 0.002 & 0.058 $\pm$ 0.002 & 0.075 $\pm$ 0.004 & 0.058 $\pm$ 0.002 \\
1.2 & 0.058 $\pm$ 0.002 & 0.058 $\pm$ 0.002 & 0.058 $\pm$ 0.002 & 0.058 $\pm$ 0.002 & 0.058 $\pm$ 0.002 & 0.058 $\pm$ 0.002 & 0.073 $\pm$ 0.004 & 0.059 $\pm$ 0.002 \\
2 & 0.061 $\pm$ 0.002 & 0.061 $\pm$ 0.002 & 0.061 $\pm$ 0.002 & 0.060 $\pm$ 0.002 & 0.060 $\pm$ 0.002 & 0.059 $\pm$ 0.002 & 0.071 $\pm$ 0.004 & 0.061 $\pm$ 0.002 \\
3 & 0.064 $\pm$ 0.003 & 0.063 $\pm$ 0.003 & 0.063 $\pm$ 0.003 & 0.063 $\pm$ 0.002 & 0.062 $\pm$ 0.002 & 0.062 $\pm$ 0.002 & 0.070 $\pm$ 0.004 & 0.064 $\pm$ 0.003 \\
4 & 0.066 $\pm$ 0.003 & 0.066 $\pm$ 0.003 & 0.065 $\pm$ 0.003 & 0.065 $\pm$ 0.003 & 0.064 $\pm$ 0.003 & 0.064 $\pm$ 0.003 & 0.070 $\pm$ 0.003 & 0.066 $\pm$ 0.003 \\
5 & 0.067 $\pm$ 0.003 & 0.067 $\pm$ 0.003 & 0.067 $\pm$ 0.003 & 0.066 $\pm$ 0.003 & 0.066 $\pm$ 0.003 & 0.065 $\pm$ 0.003 & 0.070 $\pm$ 0.003 & 0.067 $\pm$ 0.003 \\
 \bottomrule
\end{tabular}
\end{table}

\begin{table}[t]
\addtolength{\tabcolsep}{-4pt}
\centering
\caption{Classification accuracy (mean $\pm$ std) for the Body Fat dataset (Part 1).}
\begin{tabular}{c|ccccccccc}
 \toprule
\diagbox{BR}{Model} & RF[100] & RF[200] & RF[500] & RF[md10] & RF[md15] & RF[md20] & RF[md25] & RF[mss3] & RF[mss4] \\
 \midrule
0.2 & 0.057 $\pm$ 0.023 & 0.056 $\pm$ 0.023 & 0.055 $\pm$ 0.023 & 0.057 $\pm$ 0.023 & 0.057 $\pm$ 0.023 & 0.057 $\pm$ 0.023 & 0.057 $\pm$ 0.023 & 0.057 $\pm$ 0.023 & 0.058 $\pm$ 0.023 \\
0.4 & 0.041 $\pm$ 0.018 & 0.040 $\pm$ 0.018 & 0.040 $\pm$ 0.018 & 0.041 $\pm$ 0.018 & 0.041 $\pm$ 0.018 & 0.041 $\pm$ 0.018 & 0.041 $\pm$ 0.018 & 0.041 $\pm$ 0.018 & 0.042 $\pm$ 0.018 \\
0.6 & 0.038 $\pm$ 0.016 & 0.037 $\pm$ 0.016 & 0.037 $\pm$ 0.016 & 0.038 $\pm$ 0.016 & 0.038 $\pm$ 0.016 & 0.038 $\pm$ 0.016 & 0.038 $\pm$ 0.016 & 0.038 $\pm$ 0.016 & 0.038 $\pm$ 0.017 \\
0.8 & 0.037 $\pm$ 0.016 & 0.037 $\pm$ 0.016 & 0.036 $\pm$ 0.016 & 0.037 $\pm$ 0.016 & 0.037 $\pm$ 0.016 & 0.037 $\pm$ 0.016 & 0.037 $\pm$ 0.016 & 0.037 $\pm$ 0.016 & 0.038 $\pm$ 0.016 \\
1 & 0.037 $\pm$ 0.016 & 0.037 $\pm$ 0.016 & 0.036 $\pm$ 0.016 & 0.037 $\pm$ 0.016 & 0.037 $\pm$ 0.016 & 0.037 $\pm$ 0.016 & 0.037 $\pm$ 0.016 & 0.037 $\pm$ 0.016 & 0.038 $\pm$ 0.016 \\
1.2 & 0.037 $\pm$ 0.016 & 0.037 $\pm$ 0.016 & 0.037 $\pm$ 0.016 & 0.037 $\pm$ 0.016 & 0.037 $\pm$ 0.016 & 0.037 $\pm$ 0.016 & 0.037 $\pm$ 0.016 & 0.037 $\pm$ 0.016 & 0.037 $\pm$ 0.016 \\
2 & 0.038 $\pm$ 0.018 & 0.038 $\pm$ 0.017 & 0.038 $\pm$ 0.018 & 0.038 $\pm$ 0.018 & 0.038 $\pm$ 0.018 & 0.038 $\pm$ 0.018 & 0.038 $\pm$ 0.018 & 0.039 $\pm$ 0.018 & 0.039 $\pm$ 0.018 \\
\bottomrule
\end{tabular}
\end{table}

\begin{table}[t]
\addtolength{\tabcolsep}{-4pt}
\centering
\caption{Classification accuracy (mean $\pm$ std) for the Body Fat dataset (Part 2).}
\begin{tabular}{c|cccccccc}
 \toprule
\diagbox{BR}{Model} & RF[mss6] & RF[mss8] & RF[msl2] & RF[msl3] & RF[msl4] & RF[msl5] & RF[mfLog2] & RF[mfAll] \\
 \midrule
0.2 & 0.063 $\pm$ 0.026 & 0.073 $\pm$ 0.029 & 0.057 $\pm$ 0.025 & 0.065 $\pm$ 0.028 & 0.080 $\pm$ 0.032 & 0.102 $\pm$ 0.036 & 0.233 $\pm$ 0.047 & 0.057 $\pm$ 0.023 \\
0.4 & 0.043 $\pm$ 0.019 & 0.047 $\pm$ 0.020 & 0.041 $\pm$ 0.018 & 0.043 $\pm$ 0.019 & 0.048 $\pm$ 0.021 & 0.056 $\pm$ 0.024 & 0.177 $\pm$ 0.040 & 0.041 $\pm$ 0.018 \\
0.6 & 0.039 $\pm$ 0.017 & 0.041 $\pm$ 0.018 & 0.037 $\pm$ 0.016 & 0.039 $\pm$ 0.017 & 0.042 $\pm$ 0.018 & 0.046 $\pm$ 0.019 & 0.153 $\pm$ 0.036 & 0.038 $\pm$ 0.016 \\
0.8 & 0.039 $\pm$ 0.016 & 0.040 $\pm$ 0.017 & 0.036 $\pm$ 0.016 & 0.038 $\pm$ 0.016 & 0.040 $\pm$ 0.017 & 0.042 $\pm$ 0.018 & 0.139 $\pm$ 0.033 & 0.037 $\pm$ 0.016 \\
1 & 0.038 $\pm$ 0.017 & 0.040 $\pm$ 0.017 & 0.036 $\pm$ 0.015 & 0.037 $\pm$ 0.016 & 0.039 $\pm$ 0.016 & 0.041 $\pm$ 0.017 & 0.132 $\pm$ 0.033 & 0.037 $\pm$ 0.016 \\
1.2 & 0.038 $\pm$ 0.016 & 0.038 $\pm$ 0.017 & 0.035 $\pm$ 0.016 & 0.036 $\pm$ 0.015 & 0.037 $\pm$ 0.015 & 0.038 $\pm$ 0.016 & 0.128 $\pm$ 0.032 & 0.037 $\pm$ 0.016 \\
2 & 0.039 $\pm$ 0.018 & 0.039 $\pm$ 0.018 & 0.036 $\pm$ 0.016 & 0.035 $\pm$ 0.015 & 0.036 $\pm$ 0.015 & 0.037 $\pm$ 0.015 & 0.119 $\pm$ 0.032 & 0.038 $\pm$ 0.018 \\
3 & 0.040 $\pm$ 0.019 & 0.040 $\pm$ 0.019 & 0.038 $\pm$ 0.017 & 0.036 $\pm$ 0.016 & 0.036 $\pm$ 0.015 & 0.036 $\pm$ 0.015 & 0.115 $\pm$ 0.031 & 0.040 $\pm$ 0.018 \\
4 & 0.041 $\pm$ 0.019 & 0.041 $\pm$ 0.019 & 0.040 $\pm$ 0.018 & 0.037 $\pm$ 0.016 & 0.036 $\pm$ 0.015 & 0.036 $\pm$ 0.015 & 0.114 $\pm$ 0.031 & 0.041 $\pm$ 0.019 \\
5 & 0.041 $\pm$ 0.019 & 0.041 $\pm$ 0.019 & 0.040 $\pm$ 0.018 & 0.039 $\pm$ 0.017 & 0.037 $\pm$ 0.016 & 0.036 $\pm$ 0.015 & 0.113 $\pm$ 0.030 & 0.041 $\pm$ 0.019 \\
 \bottomrule
\end{tabular}
\end{table}

\begin{table}[t]
\addtolength{\tabcolsep}{-4pt}
\centering
\caption{Classification accuracy (mean $\pm$ std) for the Bigmart Sales dataset (Part 1).}
\begin{tabular}{c|ccccccccc}
 \toprule
\diagbox{BR}{Model} & RF[100] & RF[200] & RF[500] & RF[md10] & RF[md15] & RF[md20] & RF[md25] & RF[mss3] & RF[mss4] \\
 \midrule
0.2 & 0.414 $\pm$ 0.010 & 0.412 $\pm$ 0.010 & 0.410 $\pm$ 0.010 & 0.409 $\pm$ 0.010 & 0.413 $\pm$ 0.010 & 0.414 $\pm$ 0.010 & 0.414 $\pm$ 0.010 & 0.413 $\pm$ 0.010 & 0.413 $\pm$ 0.010 \\
0.4 & 0.424 $\pm$ 0.010 & 0.422 $\pm$ 0.010 & 0.421 $\pm$ 0.010 & 0.413 $\pm$ 0.010 & 0.421 $\pm$ 0.010 & 0.424 $\pm$ 0.010 & 0.424 $\pm$ 0.010 & 0.424 $\pm$ 0.010 & 0.423 $\pm$ 0.010 \\
0.6 & 0.433 $\pm$ 0.010 & 0.431 $\pm$ 0.010 & 0.429 $\pm$ 0.010 & 0.415 $\pm$ 0.010 & 0.427 $\pm$ 0.010 & 0.432 $\pm$ 0.010 & 0.433 $\pm$ 0.010 & 0.432 $\pm$ 0.010 & 0.431 $\pm$ 0.010 \\
0.8 & 0.439 $\pm$ 0.011 & 0.438 $\pm$ 0.011 & 0.436 $\pm$ 0.010 & 0.417 $\pm$ 0.010 & 0.431 $\pm$ 0.011 & 0.438 $\pm$ 0.010 & 0.439 $\pm$ 0.011 & 0.439 $\pm$ 0.011 & 0.437 $\pm$ 0.011 \\
1 & 0.446 $\pm$ 0.011 & 0.444 $\pm$ 0.010 & 0.443 $\pm$ 0.010 & 0.418 $\pm$ 0.010 & 0.435 $\pm$ 0.011 & 0.444 $\pm$ 0.011 & 0.446 $\pm$ 0.011 & 0.445 $\pm$ 0.011 & 0.444 $\pm$ 0.011 \\
1.2 & 0.451 $\pm$ 0.011 & 0.450 $\pm$ 0.011 & 0.448 $\pm$ 0.011 & 0.419 $\pm$ 0.010 & 0.439 $\pm$ 0.011 & 0.449 $\pm$ 0.011 & 0.451 $\pm$ 0.011 & 0.451 $\pm$ 0.011 & 0.450 $\pm$ 0.011 \\
2 & 0.467 $\pm$ 0.011 & 0.466 $\pm$ 0.011 & 0.465 $\pm$ 0.011 & 0.423 $\pm$ 0.010 & 0.449 $\pm$ 0.011 & 0.463 $\pm$ 0.011 & 0.467 $\pm$ 0.011 & 0.467 $\pm$ 0.011 & 0.466 $\pm$ 0.011 \\
\bottomrule
\end{tabular}
\end{table}

\begin{table}[t]
\addtolength{\tabcolsep}{-4pt}
\centering
\caption{Classification accuracy (mean $\pm$ std) for the Bigmart Sales dataset (Part 2).}
\begin{tabular}{c|cccccccc}
 \toprule
\diagbox{BR}{Model} & RF[mss6] & RF[mss8] & RF[msl2] & RF[msl3] & RF[msl4] & RF[msl5] & RF[mfLog2] & RF[mfAll] \\
 \midrule
0.2 & 0.412 $\pm$ 0.010 & 0.411 $\pm$ 0.010 & 0.411 $\pm$ 0.010 & 0.409 $\pm$ 0.010 & 0.407 $\pm$ 0.009 & 0.406 $\pm$ 0.009 & 0.450 $\pm$ 0.013 & 0.414 $\pm$ 0.010 \\
0.4 & 0.421 $\pm$ 0.010 & 0.419 $\pm$ 0.010 & 0.421 $\pm$ 0.010 & 0.417 $\pm$ 0.010 & 0.415 $\pm$ 0.010 & 0.413 $\pm$ 0.010 & 0.449 $\pm$ 0.011 & 0.424 $\pm$ 0.010 \\
0.6 & 0.429 $\pm$ 0.010 & 0.426 $\pm$ 0.010 & 0.428 $\pm$ 0.010 & 0.424 $\pm$ 0.010 & 0.420 $\pm$ 0.010 & 0.418 $\pm$ 0.010 & 0.452 $\pm$ 0.011 & 0.433 $\pm$ 0.010 \\
0.8 & 0.435 $\pm$ 0.011 & 0.432 $\pm$ 0.010 & 0.434 $\pm$ 0.010 & 0.429 $\pm$ 0.010 & 0.425 $\pm$ 0.010 & 0.421 $\pm$ 0.010 & 0.457 $\pm$ 0.012 & 0.439 $\pm$ 0.011 \\
1 & 0.440 $\pm$ 0.011 & 0.437 $\pm$ 0.010 & 0.440 $\pm$ 0.010 & 0.433 $\pm$ 0.010 & 0.429 $\pm$ 0.011 & 0.425 $\pm$ 0.010 & 0.461 $\pm$ 0.012 & 0.446 $\pm$ 0.011 \\
1.2 & 0.447 $\pm$ 0.011 & 0.444 $\pm$ 0.011 & 0.445 $\pm$ 0.011 & 0.440 $\pm$ 0.010 & 0.435 $\pm$ 0.010 & 0.431 $\pm$ 0.010 & 0.465 $\pm$ 0.011 & 0.451 $\pm$ 0.011 \\
2 & 0.463 $\pm$ 0.011 & 0.459 $\pm$ 0.011 & 0.460 $\pm$ 0.011 & 0.454 $\pm$ 0.011 & 0.449 $\pm$ 0.011 & 0.444 $\pm$ 0.010 & 0.476 $\pm$ 0.012 & 0.467 $\pm$ 0.011 \\
3 & 0.477 $\pm$ 0.011 & 0.474 $\pm$ 0.011 & 0.474 $\pm$ 0.011 & 0.468 $\pm$ 0.011 & 0.464 $\pm$ 0.011 & 0.458 $\pm$ 0.011 & 0.483 $\pm$ 0.012 & 0.480 $\pm$ 0.012 \\
4 & 0.486 $\pm$ 0.012 & 0.484 $\pm$ 0.012 & 0.484 $\pm$ 0.012 & 0.478 $\pm$ 0.012 & 0.474 $\pm$ 0.012 & 0.470 $\pm$ 0.011 & 0.487 $\pm$ 0.012 & 0.487 $\pm$ 0.012 \\
5 & 0.492 $\pm$ 0.012 & 0.491 $\pm$ 0.012 & 0.491 $\pm$ 0.012 & 0.487 $\pm$ 0.012 & 0.481 $\pm$ 0.012 & 0.478 $\pm$ 0.011 & 0.488 $\pm$ 0.012 & 0.493 $\pm$ 0.012 \\
 \bottomrule
\end{tabular}
\end{table}

\begin{table}[t]
\addtolength{\tabcolsep}{-4pt}
\centering
\caption{Classification accuracy (mean $\pm$ std) for the Beer Consumption dataset (Part 1).}
\begin{tabular}{c|ccccccccc}
 \toprule
\diagbox{BR}{Model} & RF[100] & RF[200] & RF[500] & RF[md10] & RF[md15] & RF[md20] & RF[md25] & RF[mss3] & RF[mss4] \\
 \midrule
0.2 & 0.379 $\pm$ 0.032 & 0.377 $\pm$ 0.031 & 0.375 $\pm$ 0.031 & 0.379 $\pm$ 0.032 & 0.379 $\pm$ 0.032 & 0.379 $\pm$ 0.032 & 0.379 $\pm$ 0.032 & 0.380 $\pm$ 0.032 & 0.381 $\pm$ 0.032 \\
0.4 & 0.355 $\pm$ 0.029 & 0.353 $\pm$ 0.028 & 0.352 $\pm$ 0.028 & 0.355 $\pm$ 0.029 & 0.355 $\pm$ 0.029 & 0.355 $\pm$ 0.029 & 0.355 $\pm$ 0.029 & 0.355 $\pm$ 0.029 & 0.355 $\pm$ 0.028 \\
0.6 & 0.355 $\pm$ 0.030 & 0.354 $\pm$ 0.029 & 0.353 $\pm$ 0.029 & 0.355 $\pm$ 0.030 & 0.355 $\pm$ 0.030 & 0.355 $\pm$ 0.030 & 0.355 $\pm$ 0.030 & 0.355 $\pm$ 0.030 & 0.355 $\pm$ 0.030 \\
0.8 & 0.361 $\pm$ 0.030 & 0.360 $\pm$ 0.030 & 0.358 $\pm$ 0.030 & 0.361 $\pm$ 0.030 & 0.361 $\pm$ 0.030 & 0.361 $\pm$ 0.030 & 0.361 $\pm$ 0.030 & 0.360 $\pm$ 0.031 & 0.360 $\pm$ 0.031 \\
1 & 0.367 $\pm$ 0.031 & 0.365 $\pm$ 0.031 & 0.364 $\pm$ 0.031 & 0.367 $\pm$ 0.031 & 0.367 $\pm$ 0.031 & 0.367 $\pm$ 0.031 & 0.367 $\pm$ 0.031 & 0.367 $\pm$ 0.031 & 0.366 $\pm$ 0.031 \\
1.2 & 0.372 $\pm$ 0.032 & 0.371 $\pm$ 0.032 & 0.370 $\pm$ 0.031 & 0.372 $\pm$ 0.031 & 0.372 $\pm$ 0.032 & 0.372 $\pm$ 0.032 & 0.372 $\pm$ 0.032 & 0.372 $\pm$ 0.031 & 0.372 $\pm$ 0.032 \\
2 & 0.393 $\pm$ 0.034 & 0.392 $\pm$ 0.035 & 0.391 $\pm$ 0.034 & 0.392 $\pm$ 0.034 & 0.393 $\pm$ 0.034 & 0.393 $\pm$ 0.034 & 0.393 $\pm$ 0.034 & 0.392 $\pm$ 0.034 & 0.392 $\pm$ 0.034 \\
\bottomrule
\end{tabular}
\end{table}

\begin{table}[t]
\addtolength{\tabcolsep}{-4pt}
\centering
\caption{Classification accuracy (mean $\pm$ std) for the Beer Consumption dataset (Part 2).}
\begin{tabular}{c|cccccccc}
 \toprule
\diagbox{BR}{Model} & RF[mss6] & RF[mss8] & RF[msl2] & RF[msl3] & RF[msl4] & RF[msl5] & RF[mfLog2] & RF[mfAll] \\
 \midrule
0.2 & 0.384 $\pm$ 0.033 & 0.389 $\pm$ 0.034 & 0.384 $\pm$ 0.031 & 0.395 $\pm$ 0.034 & 0.410 $\pm$ 0.037 & 0.430 $\pm$ 0.041 & 0.400 $\pm$ 0.032 & 0.379 $\pm$ 0.032 \\
0.4 & 0.355 $\pm$ 0.028 & 0.357 $\pm$ 0.029 & 0.355 $\pm$ 0.028 & 0.358 $\pm$ 0.029 & 0.363 $\pm$ 0.030 & 0.369 $\pm$ 0.031 & 0.377 $\pm$ 0.031 & 0.355 $\pm$ 0.029 \\
0.6 & 0.354 $\pm$ 0.029 & 0.353 $\pm$ 0.029 & 0.354 $\pm$ 0.030 & 0.355 $\pm$ 0.030 & 0.356 $\pm$ 0.030 & 0.360 $\pm$ 0.031 & 0.372 $\pm$ 0.030 & 0.355 $\pm$ 0.030 \\
0.8 & 0.357 $\pm$ 0.030 & 0.356 $\pm$ 0.030 & 0.357 $\pm$ 0.031 & 0.356 $\pm$ 0.031 & 0.356 $\pm$ 0.031 & 0.358 $\pm$ 0.032 & 0.373 $\pm$ 0.030 & 0.361 $\pm$ 0.030 \\
1 & 0.362 $\pm$ 0.031 & 0.360 $\pm$ 0.031 & 0.363 $\pm$ 0.032 & 0.360 $\pm$ 0.032 & 0.359 $\pm$ 0.032 & 0.359 $\pm$ 0.032 & 0.376 $\pm$ 0.030 & 0.367 $\pm$ 0.031 \\
1.2 & 0.369 $\pm$ 0.031 & 0.367 $\pm$ 0.031 & 0.368 $\pm$ 0.031 & 0.365 $\pm$ 0.031 & 0.362 $\pm$ 0.031 & 0.361 $\pm$ 0.032 & 0.377 $\pm$ 0.030 & 0.372 $\pm$ 0.032 \\
2 & 0.389 $\pm$ 0.034 & 0.386 $\pm$ 0.035 & 0.386 $\pm$ 0.035 & 0.381 $\pm$ 0.034 & 0.377 $\pm$ 0.034 & 0.373 $\pm$ 0.035 & 0.383 $\pm$ 0.030 & 0.393 $\pm$ 0.034 \\
3 & 0.408 $\pm$ 0.036 & 0.406 $\pm$ 0.035 & 0.405 $\pm$ 0.036 & 0.397 $\pm$ 0.036 & 0.393 $\pm$ 0.036 & 0.389 $\pm$ 0.036 & 0.386 $\pm$ 0.031 & 0.410 $\pm$ 0.036 \\
4 & 0.422 $\pm$ 0.037 & 0.420 $\pm$ 0.038 & 0.419 $\pm$ 0.037 & 0.411 $\pm$ 0.037 & 0.406 $\pm$ 0.037 & 0.402 $\pm$ 0.037 & 0.388 $\pm$ 0.031 & 0.423 $\pm$ 0.038 \\
5 & 0.431 $\pm$ 0.039 & 0.429 $\pm$ 0.039 & 0.429 $\pm$ 0.038 & 0.423 $\pm$ 0.038 & 0.416 $\pm$ 0.039 & 0.413 $\pm$ 0.039 & 0.388 $\pm$ 0.030 & 0.431 $\pm$ 0.039 \\
 \bottomrule
\end{tabular}
\end{table}

\begin{table}[t]
\addtolength{\tabcolsep}{-4pt}
\centering
\caption{Classification accuracy (mean $\pm$ std) for the Auction Verification dataset (Part 1).}
\begin{tabular}{c|ccccccccc}
 \toprule
\diagbox{BR}{Model} & RF[100] & RF[200] & RF[500] & RF[md10] & RF[md15] & RF[md20] & RF[md25] & RF[mss3] & RF[mss4] \\
 \midrule
0.2 & 0.049 $\pm$ 0.008 & 0.048 $\pm$ 0.008 & 0.048 $\pm$ 0.008 & 0.049 $\pm$ 0.008 & 0.049 $\pm$ 0.008 & 0.049 $\pm$ 0.008 & 0.049 $\pm$ 0.008 & 0.050 $\pm$ 0.009 & 0.053 $\pm$ 0.009 \\
0.4 & 0.030 $\pm$ 0.007 & 0.029 $\pm$ 0.007 & 0.029 $\pm$ 0.007 & 0.030 $\pm$ 0.007 & 0.030 $\pm$ 0.007 & 0.030 $\pm$ 0.007 & 0.030 $\pm$ 0.007 & 0.030 $\pm$ 0.007 & 0.032 $\pm$ 0.007 \\
0.6 & 0.021 $\pm$ 0.006 & 0.021 $\pm$ 0.006 & 0.021 $\pm$ 0.006 & 0.022 $\pm$ 0.006 & 0.021 $\pm$ 0.006 & 0.021 $\pm$ 0.006 & 0.021 $\pm$ 0.006 & 0.022 $\pm$ 0.006 & 0.023 $\pm$ 0.006 \\
0.8 & 0.017 $\pm$ 0.006 & 0.017 $\pm$ 0.006 & 0.017 $\pm$ 0.006 & 0.018 $\pm$ 0.006 & 0.017 $\pm$ 0.006 & 0.017 $\pm$ 0.006 & 0.017 $\pm$ 0.006 & 0.017 $\pm$ 0.006 & 0.018 $\pm$ 0.006 \\
1 & 0.015 $\pm$ 0.006 & 0.015 $\pm$ 0.006 & 0.015 $\pm$ 0.006 & 0.017 $\pm$ 0.006 & 0.015 $\pm$ 0.006 & 0.015 $\pm$ 0.006 & 0.015 $\pm$ 0.006 & 0.015 $\pm$ 0.006 & 0.016 $\pm$ 0.006 \\
1.2 & 0.014 $\pm$ 0.006 & 0.014 $\pm$ 0.006 & 0.014 $\pm$ 0.006 & 0.015 $\pm$ 0.006 & 0.014 $\pm$ 0.006 & 0.014 $\pm$ 0.006 & 0.014 $\pm$ 0.006 & 0.014 $\pm$ 0.006 & 0.014 $\pm$ 0.005 \\
2 & 0.013 $\pm$ 0.006 & 0.013 $\pm$ 0.006 & 0.013 $\pm$ 0.006 & 0.014 $\pm$ 0.006 & 0.013 $\pm$ 0.006 & 0.013 $\pm$ 0.006 & 0.013 $\pm$ 0.006 & 0.013 $\pm$ 0.006 & 0.013 $\pm$ 0.006 \\
\bottomrule
\end{tabular}
\end{table}

\begin{table}[t]
\addtolength{\tabcolsep}{-4pt}
\centering
\caption{Classification accuracy (mean $\pm$ std) for the Auction Verification dataset (Part 2).}
\begin{tabular}{c|cccccccc}
 \toprule
\diagbox{BR}{Model} & RF[mss6] & RF[mss8] & RF[msl2] & RF[msl3] & RF[msl4] & RF[msl5] & RF[mfLog2] & RF[mfAll] \\
 \midrule
0.2 & 0.062 $\pm$ 0.010 & 0.071 $\pm$ 0.010 & 0.060 $\pm$ 0.009 & 0.073 $\pm$ 0.010 & 0.083 $\pm$ 0.009 & 0.088 $\pm$ 0.009 & 0.095 $\pm$ 0.011 & 0.049 $\pm$ 0.008 \\
0.4 & 0.037 $\pm$ 0.008 & 0.044 $\pm$ 0.009 & 0.038 $\pm$ 0.008 & 0.048 $\pm$ 0.009 & 0.059 $\pm$ 0.011 & 0.070 $\pm$ 0.011 & 0.072 $\pm$ 0.009 & 0.030 $\pm$ 0.007 \\
0.6 & 0.026 $\pm$ 0.007 & 0.031 $\pm$ 0.008 & 0.029 $\pm$ 0.007 & 0.038 $\pm$ 0.009 & 0.045 $\pm$ 0.010 & 0.054 $\pm$ 0.012 & 0.063 $\pm$ 0.009 & 0.021 $\pm$ 0.006 \\
0.8 & 0.021 $\pm$ 0.006 & 0.025 $\pm$ 0.007 & 0.024 $\pm$ 0.006 & 0.032 $\pm$ 0.008 & 0.039 $\pm$ 0.009 & 0.045 $\pm$ 0.011 & 0.058 $\pm$ 0.008 & 0.017 $\pm$ 0.006 \\
1 & 0.018 $\pm$ 0.006 & 0.022 $\pm$ 0.006 & 0.021 $\pm$ 0.006 & 0.028 $\pm$ 0.007 & 0.036 $\pm$ 0.009 & 0.042 $\pm$ 0.010 & 0.055 $\pm$ 0.008 & 0.015 $\pm$ 0.006 \\
1.2 & 0.015 $\pm$ 0.005 & 0.017 $\pm$ 0.006 & 0.017 $\pm$ 0.006 & 0.022 $\pm$ 0.007 & 0.027 $\pm$ 0.008 & 0.033 $\pm$ 0.009 & 0.054 $\pm$ 0.008 & 0.014 $\pm$ 0.006 \\
2 & 0.012 $\pm$ 0.006 & 0.013 $\pm$ 0.006 & 0.014 $\pm$ 0.006 & 0.017 $\pm$ 0.006 & 0.020 $\pm$ 0.006 & 0.025 $\pm$ 0.007 & 0.050 $\pm$ 0.008 & 0.013 $\pm$ 0.006 \\
3 & 0.013 $\pm$ 0.006 & 0.012 $\pm$ 0.006 & 0.013 $\pm$ 0.006 & 0.014 $\pm$ 0.006 & 0.016 $\pm$ 0.006 & 0.018 $\pm$ 0.006 & 0.050 $\pm$ 0.008 & 0.013 $\pm$ 0.006 \\
4 & 0.013 $\pm$ 0.006 & 0.013 $\pm$ 0.006 & 0.013 $\pm$ 0.006 & 0.013 $\pm$ 0.006 & 0.015 $\pm$ 0.006 & 0.016 $\pm$ 0.006 & 0.050 $\pm$ 0.008 & 0.014 $\pm$ 0.006 \\
5 & 0.013 $\pm$ 0.006 & 0.013 $\pm$ 0.006 & 0.014 $\pm$ 0.006 & 0.013 $\pm$ 0.006 & 0.014 $\pm$ 0.006 & 0.015 $\pm$ 0.006 & 0.050 $\pm$ 0.008 & 0.014 $\pm$ 0.006 \\
 \bottomrule
\end{tabular}
\end{table}

\begin{table}[t]
\addtolength{\tabcolsep}{-4pt}
\centering
\caption{Classification accuracy (mean $\pm$ std) for the Advertising dataset (Part 1).}
\begin{tabular}{c|ccccccccc}
 \toprule
\diagbox{BR}{Model} & RF[100] & RF[200] & RF[500] & RF[md10] & RF[md15] & RF[md20] & RF[md25] & RF[mss3] & RF[mss4] \\
 \midrule
0.2 & 0.104 $\pm$ 0.019 & 0.103 $\pm$ 0.018 & 0.103 $\pm$ 0.018 & 0.104 $\pm$ 0.019 & 0.104 $\pm$ 0.019 & 0.104 $\pm$ 0.019 & 0.104 $\pm$ 0.019 & 0.106 $\pm$ 0.019 & 0.109 $\pm$ 0.019 \\
0.4 & 0.080 $\pm$ 0.015 & 0.079 $\pm$ 0.014 & 0.079 $\pm$ 0.014 & 0.080 $\pm$ 0.015 & 0.080 $\pm$ 0.015 & 0.080 $\pm$ 0.015 & 0.080 $\pm$ 0.015 & 0.080 $\pm$ 0.015 & 0.083 $\pm$ 0.015 \\
0.6 & 0.073 $\pm$ 0.013 & 0.072 $\pm$ 0.013 & 0.072 $\pm$ 0.013 & 0.073 $\pm$ 0.013 & 0.073 $\pm$ 0.013 & 0.073 $\pm$ 0.013 & 0.073 $\pm$ 0.013 & 0.073 $\pm$ 0.014 & 0.075 $\pm$ 0.014 \\
0.8 & 0.070 $\pm$ 0.014 & 0.070 $\pm$ 0.014 & 0.070 $\pm$ 0.013 & 0.070 $\pm$ 0.014 & 0.070 $\pm$ 0.014 & 0.070 $\pm$ 0.014 & 0.070 $\pm$ 0.014 & 0.071 $\pm$ 0.014 & 0.072 $\pm$ 0.014 \\
1 & 0.069 $\pm$ 0.014 & 0.069 $\pm$ 0.014 & 0.069 $\pm$ 0.014 & 0.069 $\pm$ 0.014 & 0.069 $\pm$ 0.014 & 0.069 $\pm$ 0.014 & 0.069 $\pm$ 0.014 & 0.070 $\pm$ 0.014 & 0.071 $\pm$ 0.014 \\
1.2 & 0.070 $\pm$ 0.014 & 0.070 $\pm$ 0.014 & 0.070 $\pm$ 0.014 & 0.070 $\pm$ 0.014 & 0.070 $\pm$ 0.014 & 0.070 $\pm$ 0.014 & 0.070 $\pm$ 0.014 & 0.070 $\pm$ 0.014 & 0.071 $\pm$ 0.015 \\
2 & 0.073 $\pm$ 0.016 & 0.072 $\pm$ 0.016 & 0.072 $\pm$ 0.016 & 0.073 $\pm$ 0.015 & 0.073 $\pm$ 0.016 & 0.073 $\pm$ 0.016 & 0.073 $\pm$ 0.016 & 0.073 $\pm$ 0.015 & 0.073 $\pm$ 0.015 \\
\bottomrule
\end{tabular}
\end{table}

\begin{table}[t]
\addtolength{\tabcolsep}{-4pt}
\centering
\caption{Classification accuracy (mean $\pm$ std) for the Advertising dataset (Part 2).}
\begin{tabular}{c|cccccccc}
 \toprule
\diagbox{BR}{Model} & RF[mss6] & RF[mss8] & RF[msl2] & RF[msl3] & RF[msl4] & RF[msl5] & RF[mfLog2] & RF[mfAll] \\
 \midrule
0.2 & 0.122 $\pm$ 0.021 & 0.148 $\pm$ 0.026 & 0.118 $\pm$ 0.022 & 0.143 $\pm$ 0.026 & 0.179 $\pm$ 0.032 & 0.224 $\pm$ 0.035 & 0.243 $\pm$ 0.045 & 0.104 $\pm$ 0.019 \\
0.4 & 0.090 $\pm$ 0.016 & 0.098 $\pm$ 0.017 & 0.086 $\pm$ 0.016 & 0.098 $\pm$ 0.017 & 0.112 $\pm$ 0.020 & 0.128 $\pm$ 0.023 & 0.178 $\pm$ 0.035 & 0.080 $\pm$ 0.015 \\
0.6 & 0.081 $\pm$ 0.015 & 0.089 $\pm$ 0.015 & 0.077 $\pm$ 0.014 & 0.086 $\pm$ 0.016 & 0.096 $\pm$ 0.017 & 0.107 $\pm$ 0.019 & 0.159 $\pm$ 0.034 & 0.073 $\pm$ 0.013 \\
0.8 & 0.077 $\pm$ 0.015 & 0.085 $\pm$ 0.016 & 0.074 $\pm$ 0.014 & 0.081 $\pm$ 0.015 & 0.091 $\pm$ 0.016 & 0.100 $\pm$ 0.017 & 0.147 $\pm$ 0.033 & 0.070 $\pm$ 0.014 \\
1 & 0.075 $\pm$ 0.015 & 0.083 $\pm$ 0.016 & 0.073 $\pm$ 0.015 & 0.079 $\pm$ 0.015 & 0.088 $\pm$ 0.017 & 0.096 $\pm$ 0.018 & 0.138 $\pm$ 0.030 & 0.069 $\pm$ 0.014 \\
1.2 & 0.073 $\pm$ 0.015 & 0.077 $\pm$ 0.015 & 0.072 $\pm$ 0.015 & 0.076 $\pm$ 0.015 & 0.080 $\pm$ 0.015 & 0.086 $\pm$ 0.017 & 0.133 $\pm$ 0.029 & 0.070 $\pm$ 0.014 \\
2 & 0.074 $\pm$ 0.015 & 0.076 $\pm$ 0.016 & 0.073 $\pm$ 0.015 & 0.076 $\pm$ 0.016 & 0.079 $\pm$ 0.016 & 0.082 $\pm$ 0.016 & 0.126 $\pm$ 0.028 & 0.073 $\pm$ 0.016 \\
3 & 0.077 $\pm$ 0.017 & 0.078 $\pm$ 0.017 & 0.076 $\pm$ 0.016 & 0.077 $\pm$ 0.017 & 0.079 $\pm$ 0.017 & 0.081 $\pm$ 0.017 & 0.124 $\pm$ 0.027 & 0.076 $\pm$ 0.017 \\
4 & 0.079 $\pm$ 0.017 & 0.079 $\pm$ 0.017 & 0.078 $\pm$ 0.017 & 0.078 $\pm$ 0.017 & 0.080 $\pm$ 0.017 & 0.081 $\pm$ 0.017 & 0.124 $\pm$ 0.028 & 0.079 $\pm$ 0.017 \\
5 & 0.080 $\pm$ 0.017 & 0.081 $\pm$ 0.017 & 0.080 $\pm$ 0.017 & 0.079 $\pm$ 0.017 & 0.080 $\pm$ 0.017 & 0.082 $\pm$ 0.017 & 0.125 $\pm$ 0.027 & 0.080 $\pm$ 0.017 \\
 \bottomrule
\end{tabular}
\end{table}

\begin{table}[t]
\addtolength{\tabcolsep}{-4pt}
\centering
\caption{Classification accuracy (mean $\pm$ std) for the Garment Productivity  dataset (Part 1).}
\begin{tabular}{c|ccccccccc}
 \toprule
\diagbox{BR}{Model} & RF[100] & RF[200] & RF[500] & RF[md10] & RF[md15] & RF[md20] & RF[md25] & RF[mss3] & RF[mss4] \\
 \midrule
0.2 & 0.560 $\pm$ 0.047 & 0.557 $\pm$ 0.046 & 0.554 $\pm$ 0.046 & 0.560 $\pm$ 0.047 & 0.560 $\pm$ 0.046 & 0.560 $\pm$ 0.047 & 0.560 $\pm$ 0.047 & 0.560 $\pm$ 0.047 & 0.561 $\pm$ 0.047 \\
0.4 & 0.534 $\pm$ 0.047 & 0.531 $\pm$ 0.047 & 0.529 $\pm$ 0.047 & 0.535 $\pm$ 0.048 & 0.534 $\pm$ 0.048 & 0.534 $\pm$ 0.047 & 0.534 $\pm$ 0.047 & 0.534 $\pm$ 0.048 & 0.534 $\pm$ 0.048 \\
0.6 & 0.530 $\pm$ 0.048 & 0.527 $\pm$ 0.047 & 0.526 $\pm$ 0.047 & 0.531 $\pm$ 0.048 & 0.530 $\pm$ 0.048 & 0.530 $\pm$ 0.048 & 0.530 $\pm$ 0.048 & 0.530 $\pm$ 0.047 & 0.529 $\pm$ 0.048 \\
0.8 & 0.534 $\pm$ 0.047 & 0.532 $\pm$ 0.047 & 0.530 $\pm$ 0.047 & 0.534 $\pm$ 0.048 & 0.534 $\pm$ 0.048 & 0.534 $\pm$ 0.048 & 0.534 $\pm$ 0.048 & 0.533 $\pm$ 0.048 & 0.532 $\pm$ 0.048 \\
1 & 0.539 $\pm$ 0.048 & 0.536 $\pm$ 0.048 & 0.535 $\pm$ 0.048 & 0.535 $\pm$ 0.049 & 0.539 $\pm$ 0.049 & 0.539 $\pm$ 0.048 & 0.539 $\pm$ 0.048 & 0.537 $\pm$ 0.049 & 0.535 $\pm$ 0.048 \\
1.2 & 0.545 $\pm$ 0.050 & 0.542 $\pm$ 0.050 & 0.541 $\pm$ 0.049 & 0.540 $\pm$ 0.050 & 0.546 $\pm$ 0.050 & 0.545 $\pm$ 0.050 & 0.545 $\pm$ 0.050 & 0.544 $\pm$ 0.050 & 0.542 $\pm$ 0.050 \\
2 & 0.570 $\pm$ 0.052 & 0.567 $\pm$ 0.051 & 0.567 $\pm$ 0.052 & 0.559 $\pm$ 0.054 & 0.570 $\pm$ 0.052 & 0.570 $\pm$ 0.052 & 0.570 $\pm$ 0.052 & 0.569 $\pm$ 0.052 & 0.566 $\pm$ 0.053 \\
\bottomrule
\end{tabular}
\end{table}

\begin{table}[t]
\addtolength{\tabcolsep}{-4pt}
\centering
\caption{Classification accuracy (mean $\pm$ std) for the Garment Productivity  dataset (Part 2).}
\begin{tabular}{c|cccccccc}
 \toprule
\diagbox{BR}{Model} & RF[mss6] & RF[mss8] & RF[msl2] & RF[msl3] & RF[msl4] & RF[msl5] & RF[mfLog2] & RF[mfAll] \\
 \midrule
0.2 & 0.563 $\pm$ 0.047 & 0.567 $\pm$ 0.048 & 0.565 $\pm$ 0.046 & 0.574 $\pm$ 0.046 & 0.582 $\pm$ 0.047 & 0.592 $\pm$ 0.048 & 0.586 $\pm$ 0.046 & 0.560 $\pm$ 0.047 \\
0.4 & 0.535 $\pm$ 0.048 & 0.537 $\pm$ 0.048 & 0.538 $\pm$ 0.047 & 0.544 $\pm$ 0.048 & 0.551 $\pm$ 0.048 & 0.558 $\pm$ 0.048 & 0.551 $\pm$ 0.043 & 0.534 $\pm$ 0.047 \\
0.6 & 0.529 $\pm$ 0.048 & 0.530 $\pm$ 0.047 & 0.531 $\pm$ 0.048 & 0.536 $\pm$ 0.048 & 0.541 $\pm$ 0.048 & 0.547 $\pm$ 0.048 & 0.540 $\pm$ 0.043 & 0.530 $\pm$ 0.048 \\
0.8 & 0.530 $\pm$ 0.048 & 0.530 $\pm$ 0.048 & 0.532 $\pm$ 0.048 & 0.534 $\pm$ 0.048 & 0.538 $\pm$ 0.047 & 0.542 $\pm$ 0.047 & 0.535 $\pm$ 0.044 & 0.534 $\pm$ 0.047 \\
1 & 0.533 $\pm$ 0.048 & 0.532 $\pm$ 0.048 & 0.534 $\pm$ 0.049 & 0.535 $\pm$ 0.048 & 0.537 $\pm$ 0.047 & 0.541 $\pm$ 0.047 & 0.532 $\pm$ 0.044 & 0.539 $\pm$ 0.048 \\
1.2 & 0.539 $\pm$ 0.049 & 0.537 $\pm$ 0.049 & 0.538 $\pm$ 0.050 & 0.537 $\pm$ 0.050 & 0.536 $\pm$ 0.049 & 0.537 $\pm$ 0.049 & 0.533 $\pm$ 0.042 & 0.545 $\pm$ 0.050 \\
2 & 0.561 $\pm$ 0.053 & 0.557 $\pm$ 0.052 & 0.557 $\pm$ 0.052 & 0.552 $\pm$ 0.053 & 0.547 $\pm$ 0.052 & 0.545 $\pm$ 0.051 & 0.535 $\pm$ 0.043 & 0.570 $\pm$ 0.052 \\
3 & 0.585 $\pm$ 0.055 & 0.580 $\pm$ 0.055 & 0.580 $\pm$ 0.055 & 0.570 $\pm$ 0.054 & 0.564 $\pm$ 0.054 & 0.560 $\pm$ 0.054 & 0.540 $\pm$ 0.043 & 0.593 $\pm$ 0.056 \\
4 & 0.603 $\pm$ 0.058 & 0.598 $\pm$ 0.058 & 0.600 $\pm$ 0.057 & 0.588 $\pm$ 0.057 & 0.580 $\pm$ 0.056 & 0.576 $\pm$ 0.056 & 0.541 $\pm$ 0.044 & 0.608 $\pm$ 0.058 \\
5 & 0.618 $\pm$ 0.061 & 0.614 $\pm$ 0.061 & 0.616 $\pm$ 0.060 & 0.605 $\pm$ 0.059 & 0.594 $\pm$ 0.058 & 0.589 $\pm$ 0.057 & 0.542 $\pm$ 0.045 & 0.620 $\pm$ 0.061 \\
 \bottomrule
\end{tabular}
\end{table}

\begin{table}[t]
\addtolength{\tabcolsep}{-4pt}
\centering
\caption{Classification accuracy (mean $\pm$ std) for the Pok mon Combat Power dataset (Part 1).}
\begin{tabular}{c|ccccccccc}
 \toprule
\diagbox{BR}{Model} & RF[100] & RF[200] & RF[500] & RF[md10] & RF[md15] & RF[md20] & RF[md25] & RF[mss3] & RF[mss4] \\
 \midrule
0.2 & 0.216 $\pm$ 0.041 & 0.213 $\pm$ 0.038 & 0.211 $\pm$ 0.038 & 0.216 $\pm$ 0.041 & 0.216 $\pm$ 0.041 & 0.216 $\pm$ 0.041 & 0.216 $\pm$ 0.041 & 0.218 $\pm$ 0.041 & 0.220 $\pm$ 0.042 \\
0.4 & 0.174 $\pm$ 0.039 & 0.171 $\pm$ 0.038 & 0.170 $\pm$ 0.037 & 0.174 $\pm$ 0.039 & 0.174 $\pm$ 0.039 & 0.174 $\pm$ 0.039 & 0.174 $\pm$ 0.039 & 0.175 $\pm$ 0.039 & 0.177 $\pm$ 0.039 \\
0.6 & 0.158 $\pm$ 0.037 & 0.156 $\pm$ 0.037 & 0.155 $\pm$ 0.037 & 0.158 $\pm$ 0.037 & 0.158 $\pm$ 0.037 & 0.158 $\pm$ 0.037 & 0.158 $\pm$ 0.037 & 0.158 $\pm$ 0.037 & 0.160 $\pm$ 0.037 \\
0.8 & 0.153 $\pm$ 0.038 & 0.151 $\pm$ 0.038 & 0.150 $\pm$ 0.037 & 0.153 $\pm$ 0.038 & 0.153 $\pm$ 0.038 & 0.153 $\pm$ 0.038 & 0.153 $\pm$ 0.038 & 0.154 $\pm$ 0.038 & 0.156 $\pm$ 0.038 \\
1 & 0.152 $\pm$ 0.040 & 0.151 $\pm$ 0.039 & 0.149 $\pm$ 0.038 & 0.152 $\pm$ 0.040 & 0.152 $\pm$ 0.040 & 0.152 $\pm$ 0.040 & 0.152 $\pm$ 0.040 & 0.152 $\pm$ 0.041 & 0.154 $\pm$ 0.040 \\
1.2 & 0.151 $\pm$ 0.040 & 0.150 $\pm$ 0.039 & 0.149 $\pm$ 0.040 & 0.151 $\pm$ 0.040 & 0.151 $\pm$ 0.040 & 0.151 $\pm$ 0.040 & 0.151 $\pm$ 0.040 & 0.152 $\pm$ 0.040 & 0.153 $\pm$ 0.040 \\
2 & 0.160 $\pm$ 0.043 & 0.159 $\pm$ 0.043 & 0.158 $\pm$ 0.043 & 0.160 $\pm$ 0.043 & 0.160 $\pm$ 0.043 & 0.160 $\pm$ 0.043 & 0.160 $\pm$ 0.043 & 0.160 $\pm$ 0.043 & 0.160 $\pm$ 0.043 \\
\bottomrule
\end{tabular}
\end{table}

\begin{table}[t]
\addtolength{\tabcolsep}{-4pt}
\centering
\caption{Classification accuracy (mean $\pm$ std) for the Pok mon Combat Power dataset (Part 2).}
%
\end{table}

\begin{table}[t]
\addtolength{\tabcolsep}{-4pt}
\centering
\caption{Classification accuracy (mean $\pm$ std) for the Auto MPG dataset (Part 1).}
\begin{tabular}{c|ccccccccc}
 \toprule
\diagbox{BR}{Model} & RF[100] & RF[200] & RF[500] & RF[md10] & RF[md15] & RF[md20] & RF[md25] & RF[mss3] & RF[mss4] \\
 \midrule
0.2 & 0.157 $\pm$ 0.022 & 0.156 $\pm$ 0.021 & 0.156 $\pm$ 0.022 & 0.157 $\pm$ 0.022 & 0.157 $\pm$ 0.022 & 0.157 $\pm$ 0.022 & 0.157 $\pm$ 0.022 & 0.158 $\pm$ 0.022 & 0.159 $\pm$ 0.022 \\
0.4 & 0.143 $\pm$ 0.020 & 0.142 $\pm$ 0.020 & 0.142 $\pm$ 0.020 & 0.143 $\pm$ 0.020 & 0.143 $\pm$ 0.020 & 0.143 $\pm$ 0.020 & 0.143 $\pm$ 0.020 & 0.143 $\pm$ 0.020 & 0.144 $\pm$ 0.020 \\
0.6 & 0.139 $\pm$ 0.019 & 0.139 $\pm$ 0.020 & 0.139 $\pm$ 0.019 & 0.139 $\pm$ 0.020 & 0.139 $\pm$ 0.019 & 0.139 $\pm$ 0.019 & 0.139 $\pm$ 0.019 & 0.140 $\pm$ 0.019 & 0.140 $\pm$ 0.020 \\
0.8 & 0.138 $\pm$ 0.020 & 0.138 $\pm$ 0.020 & 0.137 $\pm$ 0.020 & 0.138 $\pm$ 0.020 & 0.138 $\pm$ 0.020 & 0.138 $\pm$ 0.020 & 0.138 $\pm$ 0.020 & 0.138 $\pm$ 0.020 & 0.139 $\pm$ 0.020 \\
1 & 0.139 $\pm$ 0.021 & 0.138 $\pm$ 0.020 & 0.138 $\pm$ 0.020 & 0.139 $\pm$ 0.020 & 0.139 $\pm$ 0.021 & 0.139 $\pm$ 0.021 & 0.139 $\pm$ 0.021 & 0.139 $\pm$ 0.020 & 0.139 $\pm$ 0.020 \\
1.2 & 0.140 $\pm$ 0.020 & 0.139 $\pm$ 0.020 & 0.139 $\pm$ 0.020 & 0.140 $\pm$ 0.020 & 0.140 $\pm$ 0.020 & 0.140 $\pm$ 0.020 & 0.140 $\pm$ 0.020 & 0.140 $\pm$ 0.020 & 0.140 $\pm$ 0.020 \\
2 & 0.146 $\pm$ 0.021 & 0.145 $\pm$ 0.021 & 0.145 $\pm$ 0.021 & 0.146 $\pm$ 0.021 & 0.146 $\pm$ 0.021 & 0.146 $\pm$ 0.021 & 0.146 $\pm$ 0.021 & 0.146 $\pm$ 0.021 & 0.145 $\pm$ 0.021 \\
\bottomrule
\end{tabular}
\end{table}

\begin{table}[t]
\addtolength{\tabcolsep}{-4pt}
\centering
\caption{Classification accuracy (mean $\pm$ std) for the Auto MPG dataset (Part 2).}
%
\end{table}

\begin{table}[t]
\addtolength{\tabcolsep}{-4pt}
\centering
\caption{Classification accuracy (mean $\pm$ std) for the Medical Insurance dataset (Part 1).}
%
\end{table}

\begin{table}[t]
\addtolength{\tabcolsep}{-4pt}
\centering
\caption{Classification accuracy (mean $\pm$ std) for the Medical Insurance dataset (Part 2).}
%
\end{table}

\begin{table}[t]
\addtolength{\tabcolsep}{-4pt}
\centering
\caption{Classification accuracy (mean $\pm$ std) for the Laptop Price dataset (Part 1).}
%
\end{table}

\begin{table}[t]
\addtolength{\tabcolsep}{-4pt}
\centering
\caption{Classification accuracy (mean $\pm$ std) for the Laptop Price dataset (Part 2).}
%
\end{table}

\begin{table}[t]
\addtolength{\tabcolsep}{-4pt}
\centering
\caption{Classification accuracy (mean $\pm$ std) for the Household Income dataset (Part 1).}
%
\end{table}

\begin{table}[t]
\addtolength{\tabcolsep}{-4pt}
\centering
\caption{Classification accuracy (mean $\pm$ std) for the Household Income dataset (Part 2).}
%
\end{table}

\begin{table}[t]
\addtolength{\tabcolsep}{-4pt}
\centering
\caption{Classification accuracy (mean $\pm$ std) for the Liver Disorders dataset (Part 1).}
%
\end{table}

\begin{table}[t]
\addtolength{\tabcolsep}{-4pt}
\centering
\caption{Classification accuracy (mean $\pm$ std) for the Liver Disorders dataset (Part 2).}
%
\end{table}

\begin{table}[t]
\addtolength{\tabcolsep}{-4pt}
\centering
\caption{Classification accuracy (mean $\pm$ std) for the Forest Fires dataset (Part 1).}
%
\end{table}

\begin{table}[t]
\addtolength{\tabcolsep}{-4pt}
\centering
\caption{Classification accuracy (mean $\pm$ std) for the Forest Fires dataset (Part 2).}
%
\end{table}

\begin{table}[t]
\addtolength{\tabcolsep}{-4pt}
\centering
\caption{Classification accuracy (mean $\pm$ std) for the Facebook Metrics dataset (Part 1).}
%
\end{table}

\begin{table}[t]
\addtolength{\tabcolsep}{-4pt}
\centering
\caption{Classification accuracy (mean $\pm$ std) for the Facebook Metrics dataset (Part 2).}
%
\end{table}

\begin{table}[t]
\addtolength{\tabcolsep}{-4pt}
\centering
\caption{Classification accuracy (mean $\pm$ std) for the Servo dataset (Part 1).}
%
\end{table}

\begin{table}[t]
\addtolength{\tabcolsep}{-4pt}
\centering
\caption{Classification accuracy (mean $\pm$ std) for the Servo dataset (Part 2).}
%
\end{table}

\begin{table}[t]
\addtolength{\tabcolsep}{-4pt}
\centering
\caption{Classification accuracy (mean $\pm$ std) for the Concrete Compressive Strength dataset (Part 1).}
%
\end{table}

\begin{table}[t]
\addtolength{\tabcolsep}{-4pt}
\centering
\caption{Classification accuracy (mean $\pm$ std) for the Concrete Compressive Strength dataset (Part 2).}
%
\end{table}

\begin{table}[t]
\addtolength{\tabcolsep}{-4pt}
\centering
\caption{Classification accuracy (mean $\pm$ std) for the Demand Forecasting dataset (Part 1).}
%
\end{table}

\begin{table}[t]
\addtolength{\tabcolsep}{-4pt}
\centering
\caption{Classification accuracy (mean $\pm$ std) for the Demand Forecasting dataset (Part 2).}
%
\end{table}

\begin{table}[t]
\addtolength{\tabcolsep}{-4pt}
\centering
\caption{Classification accuracy (mean $\pm$ std) for the Mosquito Indicator dataset (Part 1).}
%
\end{table}

\begin{table}[t]
\addtolength{\tabcolsep}{-4pt}
\centering
\caption{Classification accuracy (mean $\pm$ std) for the Mosquito Indicator dataset (Part 2).}
%
\end{table}

\begin{table}[t]
\addtolength{\tabcolsep}{-4pt}
\centering
\caption{Classification accuracy (mean $\pm$ std) for the Brain weight dataset (Part 1).}
%
\end{table}

\begin{table}[t]
\addtolength{\tabcolsep}{-4pt}
\centering
\caption{Classification accuracy (mean $\pm$ std) for the Brain weight dataset (Part 2).}
%
\end{table}

\begin{table}[t]
\addtolength{\tabcolsep}{-4pt}
\centering
\caption{Classification accuracy (mean $\pm$ std) for the Corn Farming dataset (Part 1).}
%
\end{table}

\begin{table}[t]
\addtolength{\tabcolsep}{-4pt}
\centering
\caption{Classification accuracy (mean $\pm$ std) for the Corn Farming dataset (Part 2).}
%
\end{table}

\begin{table}[t]
\addtolength{\tabcolsep}{-4pt}
\centering
\caption{Classification accuracy (mean $\pm$ std) for the Happiness Index dataset (Part 1).}
%
\end{table}

\begin{table}[t]
\addtolength{\tabcolsep}{-4pt}
\centering
\caption{Classification accuracy (mean $\pm$ std) for the Happiness Index dataset (Part 2).}
%
\end{table}

\begin{table}[t]
\addtolength{\tabcolsep}{-4pt}
\centering
\caption{Classification accuracy (mean $\pm$ std) for the Gun Price dataset (Part 1).}
%
\end{table}

\begin{table}[t]
\addtolength{\tabcolsep}{-4pt}
\centering
\caption{Classification accuracy (mean $\pm$ std) for the Gun Price dataset (Part 2).}
%
\end{table}

\begin{table}[t]
\addtolength{\tabcolsep}{-4pt}
\centering
\caption{Classification accuracy (mean $\pm$ std) for the Fuel Consumption dataset (Part 1).}
%
\end{table}

\begin{table}[t]
\addtolength{\tabcolsep}{-4pt}
\centering
\caption{Classification accuracy (mean $\pm$ std) for the Fuel Consumption dataset (Part 2).}
%
\end{table}

\begin{table}[t]
\addtolength{\tabcolsep}{-4pt}
\centering
\caption{Classification accuracy (mean $\pm$ std) for the Motorcycle dataset (Part 1).}
%
\end{table}

\begin{table}[t]
\addtolength{\tabcolsep}{-4pt}
\centering
\caption{Classification accuracy (mean $\pm$ std) for the Motorcycle dataset (Part 2).}
%
\end{table}

\begin{table}[t]
\addtolength{\tabcolsep}{-4pt}
\centering
\caption{Classification accuracy (mean $\pm$ std) for the Toy Cement strength dataset (Part 1).}
%
\end{table}

\begin{table}[t]
\addtolength{\tabcolsep}{-4pt}
\centering
\caption{Classification accuracy (mean $\pm$ std) for the Toy Cement strength dataset (Part 2).}
%
\end{table}

\begin{table}[t]
\addtolength{\tabcolsep}{-4pt}
\centering
\caption{Classification accuracy (mean $\pm$ std) for the Infrared Thermography dataset (Part 1).}
%
\end{table}

\begin{table}[t]
\addtolength{\tabcolsep}{-4pt}
\centering
\caption{Classification accuracy (mean $\pm$ std) for the Infrared Thermography dataset (Part 2).}
%
\end{table}

\begin{table}[t]
\addtolength{\tabcolsep}{-4pt}
\centering
\caption{Classification accuracy (mean $\pm$ std) for the Fish Market dataset (Part 1).}
%
\end{table}

\begin{table}[t]
\addtolength{\tabcolsep}{-4pt}
\centering
\caption{Classification accuracy (mean $\pm$ std) for the Fish Market dataset (Part 2).}
%
\end{table}

\begin{table}[t]
\addtolength{\tabcolsep}{-4pt}
\centering
\caption{Classification accuracy (mean $\pm$ std) for the Wind Speed dataset (Part 1).}
%
\end{table}

\begin{table}[t]
\addtolength{\tabcolsep}{-4pt}
\centering
\caption{Classification accuracy (mean $\pm$ std) for the Wind Speed dataset (Part 2).}
%
\end{table}

\begin{table}[t]
\addtolength{\tabcolsep}{-4pt}
\centering
\caption{Classification accuracy (mean $\pm$ std) for the Student Performance dataset (Part 1).}
%
\end{table}

\begin{table}[t]
\addtolength{\tabcolsep}{-4pt}
\centering
\caption{Classification accuracy (mean $\pm$ std) for the Student Performance dataset (Part 2).}
%
\end{table}

\begin{table}[t]
\addtolength{\tabcolsep}{-4pt}
\centering
\caption{Classification accuracy (mean $\pm$ std) for the Child Weight at Birth dataset (Part 1).}
%
\end{table}

\begin{table}[t]
\addtolength{\tabcolsep}{-4pt}
\centering
\caption{Classification accuracy (mean $\pm$ std) for the Child Weight at Birth dataset (Part 2).}
%
\end{table}

\begin{table}[t]
\addtolength{\tabcolsep}{-4pt}
\centering
\caption{Classification accuracy (mean $\pm$ std) for the Wine Quality dataset (Part 1).}
%
\end{table}

\begin{table}[t]
\addtolength{\tabcolsep}{-4pt}
\centering
\caption{Classification accuracy (mean $\pm$ std) for the Wine Quality dataset (Part 2).}
%
\end{table}

\end{document}